\documentclass[10pt, conference, letterpaper]{IEEEtran}
\IEEEoverridecommandlockouts
\usepackage{cite}
\usepackage{amsmath,amssymb,amsfonts}
\usepackage{graphicx}
\usepackage{textcomp}
\usepackage{xcolor}
\usepackage{multirow}
\usepackage{algorithm}
\usepackage{algorithmic}
\usepackage{color}
\usepackage{threeparttable}
\usepackage{booktabs} 
\usepackage{hyperref}
\usepackage{graphicx}
\usepackage{subfigure} 
\usepackage{mathtools}
\DeclarePairedDelimiterX\set[1]\lbrace\rbrace{\def\given{\;\delimsize\vert\;}#1}

\newcommand{\traj}{\mathcal{T}}
\usepackage{ulem} 
\usepackage[binary-units]{siunitx}
\usepackage{cleveref}
\crefname{table}{Table}{Tables}
\crefname{figure}{Figure}{Figures}
\crefname{section}{Section}{Sections}

\usepackage{ifthen}
\newboolean{review}
\setboolean{review}{true} 

\ifthenelse{\boolean{review}}{
\usepackage{ulem}

\newcommand{\revdel}[1]{{\color{red}#1}}

\newcommand{\fixS}[1]{{\color{blue}\sout{#1}}}

}{
\newcommand{\revdel}[1]{{}}

\newcommand{\fixS}[1]{}

}

\usepackage[acronym]{glossaries}
\newacronym{ar}{AR}{Augmented Reality}
\newacronym{vr}{VR}{Virtual Reality}
\newacronym{mr}{MR}{Mixed Reality}
\newacronym{mse}{MSE}{Mean Squared Error}
\newacronym{drl}{DRL}{Deep Reinforcement Learning}
\newacronym{kc-td3}{KC-TD3}{Knowledge-assisted Constrained Twin-Delayed Deep Deterministic}
\newacronym{mtp}{MTP}{Motion-To-Photon}
\newacronym{embb}{eMBB}{Enhanced Mobile Broadband}
\newacronym{urllc}{URLLC}{Ultra-Reliable Low-Latency Communication}
\newacronym{mmtc}{mMTC}{Massive Machine Type Communication}
\newacronym{e2e}{E2E}{End-to-End}
\newacronym{aoi}{AoI}{Age of Information}
\newacronym{iiot}{IIoT}{Industrial Internet of Things}
\newacronym{mi}{MI}{Mutual Information}
\newacronym{lstm}{LSTM}{Long Short-Term Memory}
\newacronym{mlp}{MLP}{Multi-Layer Perception}
\newacronym{hdmi}{HDMI}{High-Definition Multimedia Interface}
\newacronym{apdo}{APDO}{accelerated primal-dual policy optimization}
\newacronym{cmdp}{CMDP}{Constrained Markov Decision Processes}
\usepackage{mathrsfs}
\def\BibTeX{{\rm B\kern-.05em{\sc i\kern-.025em b}\kern-.08em
    T\kern-.1667em\lower.7ex\hbox{E}\kern-.125emX}}

\vspace{-0.6in}  

\begin{document}
\title{Sampling, Communication, and Prediction Co-Design for Synchronizing the Real-World Device and Digital Model in Metaverse}

\author{Zhen Meng, Changyang She, Guodong Zhao, and Daniele De Martini

\thanks{Z. Meng and G. Zhao are with James Watt School of Engineering, University of Glasgow, UK. (e-mail: z.meng.1@glasgow.ac.uk; guodong.zhao@glasgow.ac.uk)}
\thanks{C. She is with the School of Electrical and Information Engineering, University of Sydney, Australia. (e-mail: shechangyang@gmail.com)}
\thanks{D. De Martini is with Oxford Robotics Institute, University of Oxford, UK. (e-mail: daniele@robots.ox.ac.uk)}
\thanks{Corresponding author: Changyang She.}
}
\maketitle \pagestyle{empty} \thispagestyle{empty}

\begin{abstract}
The metaverse has the potential to revolutionize the next generation of the Internet by supporting highly interactive services with the help of \gls{mr} technologies; still, to provide a satisfactory experience for users, the synchronization between the physical world and its digital models is crucial.
This work proposes a sampling, communication and prediction co-design framework to minimize the communication load subject to a constraint on tracking the \gls{mse} between a real-world device and its digital model in the metaverse.
To optimize the sampling rate and the prediction horizon, we exploit expert knowledge and develop a constrained \gls{drl} algorithm, named \gls{kc-td3} policy gradient algorithm. We validate our framework on a prototype composed of a real-world robotic arm and its digital model. Compared with existing approaches: (1) When the tracking error constraint is stringent ($\text{MSE}=0.002^{\circ}$), our policy degenerates into the policy in the sampling-communication co-design framework. (2) When the tracking error constraint is mild ($\text{MSE}=0.007^{\circ}$), our policy degenerates into the policy in the prediction-communication co-design framework. (3) Our framework achieves a better trade-off between the average \gls{mse} and the average communication load compared with a communication system without sampling and prediction. For example, the average communication load can be reduced up to $87\%$ when the average track error constraint is 0.002$^{\circ}$. (4) Our policy outperforms the benchmark with the static sampling rate and prediction horizon optimized by exhaustive search, in terms of the tail probability of the tracking error. Furthermore, with the assistance of expert knowledge, the proposed algorithm \gls{kc-td3} achieves better convergence time, stability, and final policy performance.

\end{abstract}
  

\glsresetall


\section{Introduction}
Facilitated by the rapid development of \gls{ar}, \gls{vr} and \gls{mr}, the metaverse is expected to change our daily lives in different aspects, such as shopping, social interaction, healthcare~\cite{holloway2012virtual}, education~\cite{collins2008looking} and gaming~\cite{bardzell2007video}.
One of the main challenges is providing an immersive and highly interactive experience~\cite{choi2019effects}, which requires synchronization between the physical and virtual worlds to ensure smooth user motion tracking and timely feedback.
Indeed, poor synchronization can lead to chaotic interactions and dizziness~\cite{583063}. Besides, in mission-critical applications assisted by the metaverse, even slight out-of-synchronization between a real-world device and the digital model may cause serious consequences~\cite{laaki2019prototyping}.

\begin{figure}
\centering
\includegraphics[scale=0.4]{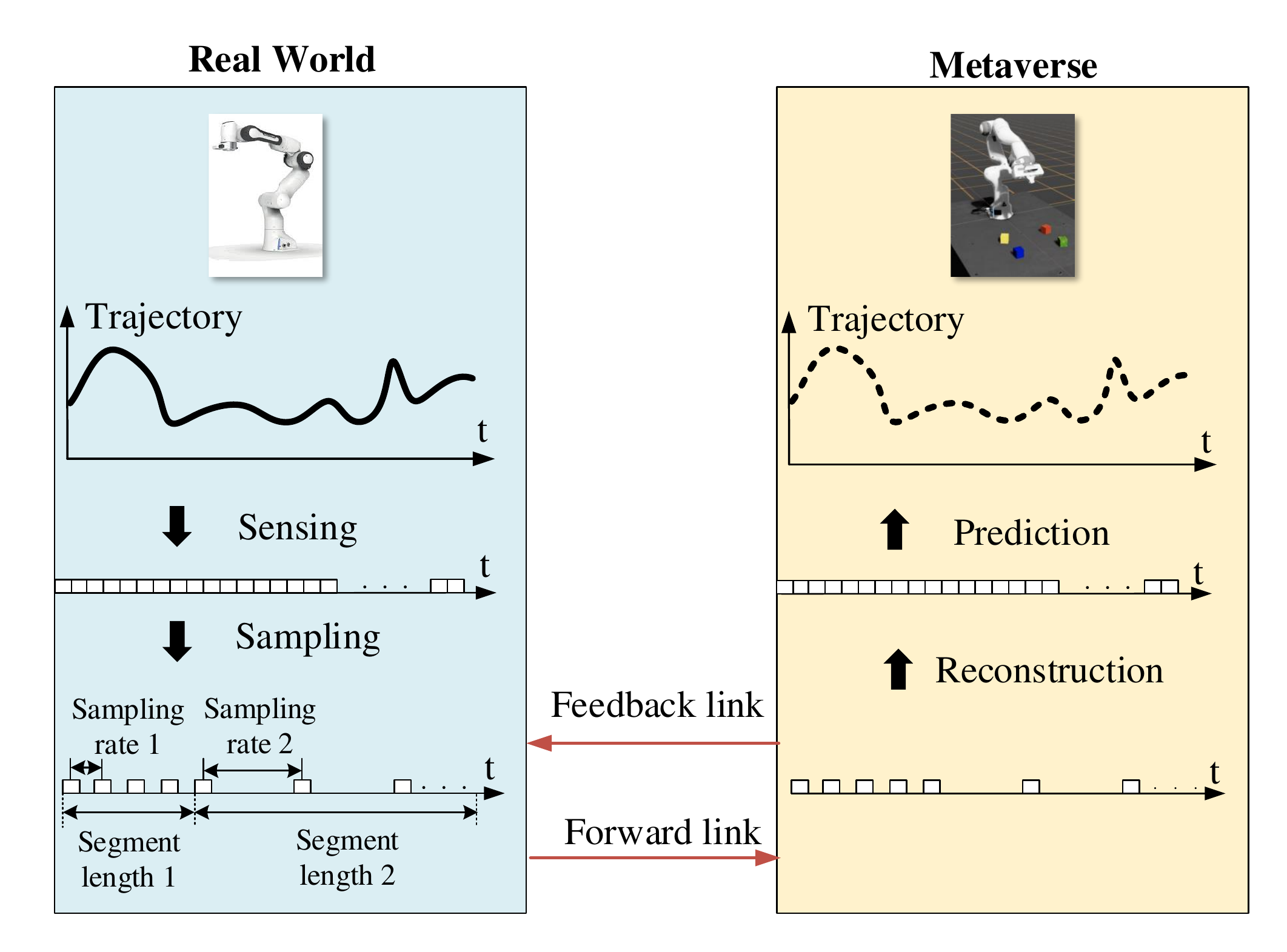}
\caption{Proposed co-design framework to synchronize a real-world device and its digital model in the metaverse.}
\label{Illustration of system model}
\end{figure}

The synchronization performance can be characterized by three key metrics: \gls{mtp} latency, data rate, and packet-loss rate.
\Gls{mtp} latency refers to the time between a user's action and the corresponding effect displayed in the virtual world~\cite{mania2004perceptual}.
In applications that require haptic feedback, the required \gls{mtp} should be less than \SI{1}{\milli\second}~\cite{7593456}.
The data rate, i.e. the maximum rate of data transfer across the network, is the bottleneck for most multimedia applications~\cite{oliver2012mongoose}, especially when a large amount of data is generated by multimedia sources -- e.g. High-Definition video, \gls{ar}/\gls{vr}/\gls{mr}, gaming, massive sensor networks, etc.
Further, the packet-loss rate, i.e. the percentage of packets sent by a transmitter but not received by the receiver, is crucial for mission-critical applications, such as remote robotic control, smart-factories monitoring and online healthcare~\cite{dianatfar2021review}. 

It is very challenging to meet the three key performance metrics in practice. Although the three use cases have been considered the fifth-generation (5G) cellular networks, \gls{embb}, \gls{urllc} and \gls{mmtc}, the requirements in metaverse cannot be fulfilled; indeed, while 5G New Radio would allow a sub-millisecond delay in the radio access network, the \gls{e2e} delay is still far from the \gls{mtp} requirement~\cite{lee2021all}.
Moreover, to support interactive applications globally, the metaverse requires an extremely high data rate, far beyond the capabilities of 5G networks at 50 Gbits per second~\cite{3GPP}.


Recently, researchers started to investigate interdisciplinary approaches beyond conventional communication system designs.
The existing literature has two branches of related work: sampling-communication co-design~\cite{wang2020reinforcement, tang2020minimizing,fountoulakis2021joint, kountouris2021semantics, pappas2021goal, 8445873} and prediction-communication co-design~\cite{hou2020motion, richter2019augmented, tong2018minimizing, hou2019prediction}.
(1) {Sampling-Communication Co-Design} uses \gls{aoi}, mutual information, or goal-oriented semantics to down-sample the information/packets at the transmitter side; thus, the task is completed with less communication load after reconstructing the original information at the receiver side.
(2) {Prediction-Communication Co-Design}, instead, predicts a device's future state based on its historical states; if the prediction horizon equals the communication delay, the user's experienced delay is zero~\cite{hou2021intelligent}.
Further, if the communication presents packet losses, the same approach can infer the missing states.

Considering that sampling, communication, and prediction are closely interconnected and interdependent, we argue that it is possible to improve the performance compared with the above existing work by combining the three.
For example, suppose that a predictor can estimate missing information in a longer prediction horizon; in that case, the overall reliability is less sensitive to packet losses in communications, and the sampling rate can be reduced.
Nevertheless, prediction and reconstruction errors may deteriorate the user experience; still, the impact on the synchronization in the metaverse remains unclear, and a coherent design framework that combines sampling, communication and prediction is not available in the existing literature.

\subsection{Related Work}
We discuss both sampling-communication and prediction-communication co-designs as present in the existing literature.  

\subsubsection{Sampling-Communication Co-design}
\Gls{aoi} is a performance metric widely used in co-design communication systems and sampling policies (also called state update policies)~\cite{wang2020reinforcement, tang2020minimizing,fountoulakis2021joint}.
In~\cite{wang2020reinforcement}, the authors optimized the sensing and updating policy for an air pollution monitoring application by minimizing the weighted sum of the \gls{aoi} and the total energy consumption of the device.
By adjusting the weighting coefficients of the \gls{aoi} and tuning the energy consumption manually, it is possible to achieve the target trade-off.
To collect new data from power-constrained sensors in an \gls{iiot} network, the authors of~\cite{tang2020minimizing} optimized a scheduling algorithm by decoupling the multi-sensor problem into single-sensor problems.
In~\cite{fountoulakis2021joint}, the authors considered a status update problem over an error-prone wireless channel, where the average cost of sampling and communications was minimized subject to average \gls{aoi} constraints.

Since \gls{aoi} is an intermediate performance metric that does not fully capture the requirement of a specific task and is not aware of the burstiness of the source, it is not a good metric \cite{kountouris2021semantics}. For example, when the state of an environment or device is stationary, there is no need to update the state frequently. When the state changes rapidly, the source should generate packets and update frequently.
For this reason, different performance metrics and design frameworks have been considered to improve sampling efficiency, e.g. goal-oriented communications~\cite{pappas2021goal} and mutual information~\cite{8445873}.
The authors of \cite{pappas2021goal} developed a goal-oriented sampling and communication policy for status updates over an unreliable wireless channel, where only effective samplings for lowing real-time reconstruction errors were allowed to be transmitted to the actuator. The results show that the proposed strategy can significantly improve the effective updates and reduce the cost of actuation errors. \cite{8445873}, instead, used the mutual information between the real-time source values and the samples delivered to the receiver to optimize the sampling policy, proposing a transmitter that maximizes the expected mutual information by sending a new packet once the latter is below a threshold.

\subsubsection{Prediction-Communication Co-design}
Prediction plays an essential role in reducing the user-experienced delay in \gls{urllc}.
To reduce round-trip delay in a \gls{vr} application, the authors of \cite{hou2020motion} proposed to predict, pre-render and cache \gls{vr} videos in an edge server, where \gls{lstm} and \gls{mlp} neural networks predict body and head motion, respectively.
In~\cite{richter2019augmented}, the authors considered an \gls{ar} robotic telesurgical application, where, with the help of prediction, they could reduce the task completion time by $19\%$ without increasing the manipulation error rate.
The authors of~\cite{tong2018minimizing} jointly optimized the communication and packetized predictive control system to minimize the wireless resource consumption under the control outage probability constraint.
The results in~\cite{hou2019prediction} indicated that prediction and communication co-design can achieve a better trade-off between reliability and latency than traditional communication systems without prediction.

\subsection{Contributions}
In this paper, we investigate how to synchronize the trajectories of a real-world device and its digital model in the metaverse. The main contributions of this paper are summarized as follows:

\begin{itemize}
\item We establish a sampling, prediction, and communication co-design framework for synchronizing the trajectories of a device in the real world and its digital model in the metaverse.
The sampling rate and the prediction horizon are jointly optimized to minimize the communication load subject to a \gls{mse} constraint between the trajectory of the device and its digital model.

\item In the co-design framework, we propose a \gls{kc-td3} algorithm by combining \gls{drl} techniques with expert knowledge on sampling, communication and prediction. Specifically, the following learning techniques are applied to improve the reinforcement learning algorithm: 1) extension of double Q-Learning, 2) state-space reduction, 3) interdependent action normalization, and 4) \gls{apdo}.

\item We build a prototype comprising a real-world robotic arm and its digital model in the metaverse.
The experimental results show that our proposed algorithm achieves good convergence time and stability.
Compared with a communication system without sampling and prediction, the sampling, communication, and prediction co-design framework can reduce the average tracking error and the communication load by \SI{87.5}{\percent} and \SI{87}{\percent}, respectively. Besides, the co-design framework works well in communication systems with high packet loss probabilities, \SIrange{1}{10}{\percent}.
\end{itemize}

The rest of this paper is organized as follows.
In \cref{sec:method}, we propose the co-design framework and formulate a joint design problem that includes sampling, communication, and prediction components in one optimization problem.
In \cref{sec:kctd3}, we develop the \gls{kc-td3} algorithm to optimize the sampling and prediction policy while minimizing the communication load.
\Cref{sec:experiment} describes the prototype we used to verify our method and \cref{sec:results} provides performance evaluations.
Finally, \cref{sec:conclusions} concludes the paper.


\section{Sampling, Communication, and Prediction Co-design Framework}
\label{sec:method}

In this section, we describe the proposed framework and formulate the co-design problem as the foundation of our algorithm.


\subsection{Co-Design Framework}
As shown in~\cref{Illustration of system model}, we consider the synchronization between a real-world device and its digital model in the metaverse. Specifically, a sensor of the device first measures its trajectory\footnote{The trajectory could be any observations such as pressure, temperature, humidity, etc. The reason why we do not use ``state" in the system design is that the definition of ``state" in our reinforcement learning algorithm is not the same as ``state" (trajectory) measured by sensors. To avoid misunderstanding, we use ``trajectory" in our system design.}, Then, the data is sampled, i.e. decimated, and transmitted to the metaverse, where it can be reconstructed and used to predict the future trajectory. To reduce the latency between the real-world devices and the digital model in the metaverse, the digital model of the device follows the predicted trajectory and feeds back the prediction results to the real-world devices. Finally, the device compares its trajectory with the predicted one and adjusts the sampling rate and the prediction horizon.

\begin{figure*}
\centering
\includegraphics[scale=0.56]{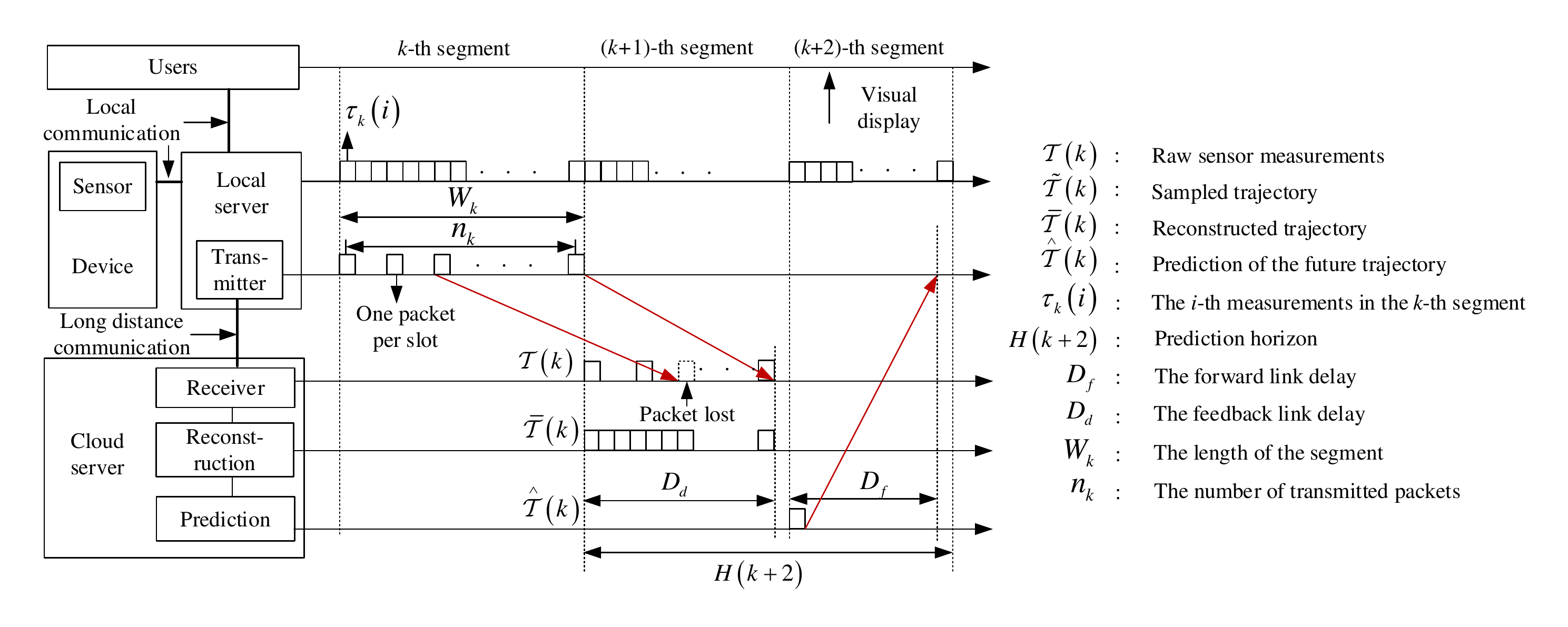}
\caption{The timing sequence of the proposed co-design framework (The senor belongs to the real world device for data generation and the transmitter belongs to a local server. The receiver and functions for construction and predication are deployed at the cloud server that operates the metaverse).}
\label{time sequence}
\end{figure*}

As shown in~\cref{time sequence}, time is discretized into slots.
The \gls{e2e} delays in the forward and feedback links of long distance communication networks are denoted by $D_d$ and $D_f$, respectively. The unit is defined as the number of time slots, where the duration of each time slot is set to 1~ms in our experiments. We assume that E2E delay includes both computing delay and communication delay. If the delay of a packet is longer than a delay bound $D_{\text{max}}$, the packet is outdated and will be dropped by the system.

The trajectory of the device is measured by the sensor in each time slot.
We can subdivide the raw sensor measurements into sequences, which we call \textit{segments}; we denote the $k$-th segment by $\traj(k) = \set{\tau_k(i) \given i = 1,\dots W_k }$, where $\tau_k(i)$ are the sensor readings and $W_k$ is the length of the segment $\traj(k)$.
We assume that the length of each segment is larger than the communication delay. As such, the $k$-th segment can arrive at the receiver by the end of the $(k+1)$-th segment. Thus, we have the following constraint
\begin{align}\label{2}
{W_k} \ge {D_{\max }}.
\end{align}

Since the raw trajectory samples are highly correlated in the time domain -- and thus the data packets -- they are highly redundant.
For this reason, to reduce the communication load, we can subsample the raw data from the sensor and then transmit them to the cloud server via the long distance communication network. Each trajectory segment is sampled at a fixed rate $r(k)$.
Let $\tilde{\traj}(k) = \set{\tilde{\tau}_k(i) \given i = 1,\dots n_k}$ be the sampled trajectories of the $k$-th segment, where $n_k$ is the number of packets sent by the transmitter.


After reception, the cloud server reconstructs the original trajectory from the sampled one -- we denote it by $\bar{\traj}(k) = \set{\bar{\tau}_k(i) \given i=1,\dots W_k }$ -- and, given the last historical trajectories $\bar{\traj}(1), \bar{\traj}(2), \dots \bar{\traj}(k)$, predicts the future trajectory of the device, i.e. $\hat{\traj}(k+1)$ and $\hat{\traj}(k+2)$ of length $W_{k+1}$ and $W_{k+2}$, respectively. As such, the prediction horizon needs to be
\begin{align}\label{re}
{H(k+2)} = {W_{k+1}}+{W_{k+2}}. 
\end{align}
Considering that the communication delay in the feedback link does not exceed $D_{\text{max}}$, which is shorter than $W_{k+2}$, the device obtains $\hat{\traj}(k+2)$ by the end of the $(k+2)$-th segment, ${\traj}(k+2)$. By measuring the average tracking error, which is defined as the \gls{mse} between the true trajectory, ${\traj}(k+2)$, and the predicted trajectory, $\hat{\traj}(k+2)$, the system can adjust the sampling rate and the length of the $(k+3)$-th segment $W_{k+3}$. 
According to (\ref{re}), in our optimization framework, instead of optimizing prediction horizon directly, we can optimize the length of each segment. Besides, the sensing data of the real-world device are displayed to the user by different equipment (e.g., High-Definition Screen, AR, VR glasses) via local communication networks (e.g., \gls{hdmi} or other outputs ports).



\subsection{Sampling, Reconstruction, and Prediction}
\subsubsection{Sampling} The transmitter only sends the sampled packets to the receiver, residing on the remote server.
Denoting the number of samples of the $k$-th segment by $n_k$, the sampling rate of the $k$-th segment is given by
\begin{align}
{r(k)} = \frac{{{n_k}}}{{{W_k}}}.\label{r}
\end{align}
As a result, the sampled trajectory can be obtained as follows,
\begin{align}\label{eq:sample}
\tilde \tau_k \left( i \right) = \tau_k \left( {\left\lfloor \frac{W_k}{n_k} \right\rfloor \cdot \left( {i - 1} \right) + 1} \right),{\rm{ }}i = 1,...,{n_k}.
\end{align}
where $\lfloor \cdot \rfloor$ represents the floor function.

\subsubsection{Reconstruction} In digital signal processing, there are several up-sampling methods for state reconstruction, such as inverse fast Fourier transform, repetitions of the last received state, linear interpolation, and barycentric interpolation~\cite{steffensen2006interpolation}.
Without loss of generality, due to its wide adoption, and its simplicity, we adopt linear interpolation~\cite{scaglia2010linear}.
The relationship between the reconstructed trajectory and the sampled trajectory is given by
\begin{align}\label{reconstruct}
\bar{\traj}(k) = f_{\rm I}(\tilde{\traj}(k);\theta_{\rm I}),
\end{align}
where $\theta_{\rm I}$ are the parameters for the interpolation function.

\subsubsection{Prediction} We utilize a \gls{mlp} that takes the historical trajectory as its input and generates the predicted one.
The relationship between the input and output of the predictor in the $t_{k+1}$-th slot can be expressed as
\begin{align}\label{predict}
& [\hat{\tau}_{k+1}  \left( t_{k+1}-W_{k+1}+1 \right),...,\hat{\tau}_{k+1} \left( {t_{k+1} + {W_{k + 2}}} \right)] \notag\\
&= f_{\rm P}(\bar \tau\left({t_{k+1} - L_{\rm{in}} - {W_{k + 1}}} \right),\bar \tau\left({t_{k+1} - L_{\rm{in}} - {W_{k + 1}}} +1\right),\notag\\
&\quad\ ...,\bar \tau \left( {t_{k+1} - {W_{k + 1}}} \right);{\theta _{\rm P}}),
\end{align}
where $L_{\rm in}$ is the length of the historical trajectory and $\theta_{\rm P}$ represents the MLP parameters. We ignore the subscript $k$ of the $\bar \tau$ since the input of the predictor includes multiple trajectory segments. It is worth noting that the value of $L_{\rm in}$ is determined by the temporal correlation of the trajectory and does not depend on the length of the $k$-th segment, $W_k$.





\subsection{Tracking Error and Communication Load}
\subsubsection{Tracking Error} The tracking error of the $k$-th segment, denoted by $e(k)$, is given by
\begin{align}\label{error}
{e(k)}&= \text{MSE}({\mathcal{T}}(k),\hat{\mathcal{T}}(k))\notag \\
&=\frac{1}{{\left| {{W_k}} \right|}}\sum\limits_{t = 1}^{{W_k}} {{{\left( {{\tau _k}(i) - {{\hat \tau}_k}(i)} \right)}^2}},
\end{align}
where $\tau_k(i)$ is the trajectory measured at the $i$-th time slot in the $k$-th segment, and ${{{\hat \tau}_k}(i)}, \text{for}~i=1,...,W_k,$ is the prediction trajectory. 


\subsubsection{Communication Load} We assume that the communication load is proportional to the packet rate of the system. For example, in Orthogonal Frequency-Division Multiplexing (OFDM) communication systems (adopted in the 4-th generation and 5-th generation cellular networks), the time-frequency resource blocks occupied by a packet are determined by the packet size and the channel gain. If the packet size and the channel gain are stationary, the average resource blocks allocated to a device are proportional to the packet rate (i.e., the sampling rate in our system). Since the number of packets to be transmitted for the $k$-th segment is $n_k$, the required radio resource blocks to transmit the $k$-th segment of the trajectory are proportional to $n_k$. Therefore, to minimize the required radio resource blocks for this service, we minimize the sampling rate subject to a constraint on the MSE in~(\ref{error}). It is also worth noting that the sampling frequency in our system is different from the sampling frequency of OFDM systems. We assume that only one packet can be transmitted in one slot (i.e., transmission time interval in OFDM systems). By adjusting the sampling rate, we adjust the packet rate in the communication system. For example, if the sampling rate is $0.1$ per slot, the transmitter only sends one packet every ten slots and remains silence in the other nine slots.

\subsection{Problem Formulation}
To reduce the communication load subject to the MSE constraint, we optimize the sampling rate and the prediction horizon. The problem can be formulated as follows,

\begin{align}
    &\min \limits_{{H(k)},n_k}\mathop {\lim }\limits_{N \to +\infty} \frac{1}{N}\sum\limits_{k = 1}^N {{r(k)}}\label{YY}\\
    & \; \text{s.t.} \notag \quad \ \;\mathop {\lim }\limits_{N \to +\infty} \frac{1}{N}\sum\limits_{k = 1}^N {{e(k)}}  \le {\Gamma _c}\tag{\ref{YY}{a}}\label{YYa},
\end{align}
where $N$ represents the number of segments and ${\Gamma _c}$ is the required average MSE threshold; this latter is dependant on specific applications and is considered fixed for both training and deployment. 

\section{Constrained Deep Reinforcement Learning for Optimizing Sampling, Prediction, and Communication}\label{Sec:CMDP}
\label{sec:kctd3}

The problem \eqref{YY} is a sequential decision problem and can be re-formulated as a \gls{cmdp}. Typically, the weighted sum reward is used for handling multi-criterion optimization (sampling rate and MSE). However, for the weighted sum reward, we need to adjust the weighting coefficient manually. With the constraint reinforcement learning, we can find the weighting coefficient by optimizing the Lagrangian multiplier in the dual domain. In other words, the constraint reinforcement learning algorithm can obtain a suitable weighting coefficient that can guarantee the MSE constraint. Therefore, to solve the problem, we integrate expert knowledge into the primal-dual Deep Deterministic Policy Gradient (DDPG) algorithm and develop the KC-TD3 algorithm.


\subsection{\gls{cmdp} Formulation}\label{Subsec:CMDP}

\textbf{State:} The state ${{\bf{s}}_k}$ observed by the device by the end of the $k$-th segment (the $t_k$-th time slot) includes the last two trajectory segments measured by the device, ${\mathcal{T}}(k-1)$ and ${\mathcal{T}}(k)$, and those predicted by the cloud sever, $\hat{\mathcal{T}}(k-1)$ and $\hat{\mathcal{T}}(k)$. The predicted trajectory segments depend on the reconstructed historical trajectory $\bar{\mathcal{T}}_{in}(k) = \bar{\tau}(t_{k}-L_{\rm in} -W_{k}-W_{k-1}),..., \bar{\tau}(t_{k}-W_{k}-W_{k-1})$. As shown in \eqref{eq:sample} and \eqref{reconstruct}, $\bar{\mathcal{T}}_{in}(k)$ is determined by the true trajectory from the $(t_{k}-L_{\rm in} -W_{k}-W_{k-1}$)-th slot to the $(t_{k}-W_{k}-W_{k-1})$-th slot, as well as the sampling rate and the prediction horizon.

\textbf{Action:} The action to be taken by the end of the $k$-th segment includes the length of the $(k+1)$-th segment, $W_{k+1}$, and the number of samples to be transmitted $n_{k+1}$ (equivalent to the sampling rate $r_{k+1})$. Thus, the action is denoted by ${{\bf{a}}_k}=(W_{k+1}, n_{k+1})$. For convenience, we denote two actions as a vector denoted by ${{\bf{a}}_k}=[{{\bf{a}}_k}^{[1]},{{\bf{a}}_k}^{[2]}]$, where ${{\bf{a}}_k}^{[1]} = W_{k+1}$ and ${{\bf{a}}_k}^{[2]}=n_{k+1}$. Based on this definition, we have ${{\bf{a}}_k}^{[1]} \in \{D_{\text{max}},...,W_{\text{max}}\}$ and ${{\bf{a}}_k}^{[2]} = {1,...,{{\bf{a}}_k}^{[1]}}$.

\textbf{Instantaneous Reward and Cost:}
Given the state and the action at the end of the $k$-th segment, the instantaneous reward is the negative of sampling rate and the cost is the MSE of the $(k+1)$-th segment. According to (\ref{r}) and (\ref{error}), we have ${r_k} = -r(k+1)$ and $c_k = e(k+1)$.

\textbf{Policy:} The agent follows a deterministic policy denoted by $\mu: {{\bf{a}}_k} = \mu \left( {{{\bf{s}}_k}|{\theta }} \right)$, where $\theta$ represents the parameters of the policy.

\textbf{Long-Term Reward and Long-Term Cost:} Following the policy $\mu \left( {{\cdot}|{\theta}} \right)$, the long-term discounted reward is given by 
\begin{align}\label{long reward}
R^{\mu \left( {{\cdot}|{\theta }} \right)} = {\mathop{\mathbb{E}}}[\sum\nolimits_{k = 0}^\infty  {{\gamma ^k}} r_k],
\end{align}
where the $\gamma$ is the discount factor. Similarly, the long-term discounted cost under the policy ${\mu(\cdot|\theta)}$ is given by
\begin{align}
C^{\mu \left( {{\cdot}|{\theta}} \right)} = {\mathop{\mathbb{E}}}[\sum\nolimits_{k = 0}^\infty   {{\gamma ^k}} {c}_k].
\end{align}

\textbf{\gls{cmdp} Formulation:} The goal is to find the optimal policy $\mu ^ *(\cdot|\theta^{*})$ that maximizes the long-term reward $R^{{\mu(\cdot|\theta)}}$ subject to the constraint on the long-term cost $C^{{\mu(\cdot|\theta)}}$. Thus, the problem can be reformulated as follows:
\begin{align}
  {\mu ^ *(\cdot|\theta^{*})} = &\arg \mathop {\max }\limits_{\mu \left( { \cdot |{\theta}} \right)} R^{{\mu(\cdot|\theta)}} \hfill \label{kk}\\
  &\,\text{s.t.}
\quad \ \; C^{{\mu(\cdot|\theta)}} \leqslant {\frac{{{\Gamma _c}}}{{1 - \gamma }}} \hfill \tag{\ref{kk}{a}} \label{kka}.
\end{align}
To utilize DRL to solve the problem, we first prove that the transitions of the system follow an MDP (see the proof in Appendix A).

The Lagrangian function of the constrained optimization problem is defined as follows \cite{bertsekas1997nonlinear},
\begin{align}\label{lagrangian}
\Gamma ({\mu(\cdot|\theta)} ,\lambda ) = R^{{\mu(\cdot|\theta)}} - \lambda (C^{{\mu(\cdot|\theta)}} - \frac{{{\Gamma _c}}}{{1 - \gamma }}),
\end{align}
where $\lambda$ is the Lagrangian multiplier. Then, the constrained problem can be converted to the following unconstrained problem,
\begin{align}
({\mu ^ *(\cdot|\theta^{*})},{\lambda ^ * }) = \arg \mathop {\min }\limits_{\lambda  \geqslant 0} \mathop {\max }\limits_{\mu \left( { \cdot |{\theta}} \right)} \Gamma ({\mu(\cdot|\theta)} ,\lambda ).
\end{align}

\subsection{Preliminary of Primal-Dual DDPG}
Primal-dual DDPG is an off-policy method to solve \gls{cmdp}~\cite{liang2018accelerated}. It combines the DDPG algorithm with the primal-dual method to find the optimal policy and the dual variable. The policy, the long-term reward, and the long-term cost are represented by three neural networks, and we denote them by $\mu \left( { \cdot |\theta } \right)$, ${Q^R}\left( { \cdot |\phi^R } \right)$ and ${Q^C}\left( { \cdot |\phi^C } \right)$, where $\theta$, $\phi^R$, and $\phi^C$ are the parameters of these three neural networks, respectively.

During the training process, the corresponding action generated by the actor network is given by
\begin{align}\label{14}
{{\bf{a}}_k} = clip\left( {{{{\mu \left( {{{\bf{s}}_k}|{\theta}} \right)}}} + clip(\varepsilon , - c,c),{\bf{a_{\rm Low}}},{\bf{a_{\rm High}}}} \right),
\end{align}
where $\varepsilon$ is the White Gaussian Noise with distribution $\varepsilon  \sim {\rm{{\cal N}}}\left( {0,\sigma } \right)$, $[\bf{a_{\rm Low}},{\bf{a_{\rm High}}}]$ is the action space, and
\begin{align}
clip(x,{c_1},{c_2}) = \min \left( {\max \left( {x,{c_1}} \right),{c_2}} \right).
\end{align}
After taking action ${{\bf{a}}_k}$ at the $k$-th step (i.e., selecting $W_k$ and $n_k$ in the last time slot of the $k$-th trajectory segment in our system), the system observes the instantaneous reward and cost, and transits from ${{\bf{s}}_k}$ to ${{\bf{s}}_{k+1}}$. The transition is denoted by ${\rm{{\cal D}}}_k \buildrel \Delta \over = \left\langle {{{\bf{s}}_k},{{\bf{a}}_k},{r_k},{c_k},{{\bf{s}}_{k + 1}}} \right\rangle$, which is stored in the replay memory, ${\rm{{\cal M}}}$. In each training step, a number of transitions (mini-batch) are randomly selected from replay memory ${\rm{{\cal M}}}$ and are used to optimize $\theta$, $\phi^R$, and $\phi^C$. We denote ${{\rm{{\cal D}}}_{{k_i}}}  = \left\langle {{{\bf{s}}_{{k_i}}},{{\bf{a}}_{{k_i}}},{r_{{k_i}}},{c_{{k_i}}},{{\bf{s}}_{{k_i} + 1}}} \right\rangle, i = 1,2,...,N_{\text{batch}}$ as the $i$-th transition in the $k$-th episode of the training stage, where $N_{\rm batch}$ is the batch size. 
To optimize the critic and cost neural networks, the Bellman equation is utilized. The target reward and cost functions can be expressed as follows:

\begin{align}\label{15}
Q^R\left( {{{\bf{s}}_{{k_i}}},{{\bf{a}}_{{k_i}}}} \right) = r + \gamma Q^R\left( {{{\bf{s}}_{{k_i} + 1}},\mu \left( {{{\bf{s}}_{{k_i} + 1}}|{\theta }} \right)}|{\phi^R} \right),
\end{align} 
\begin{align}\label{16}
Q^C\left( {{{\bf{s}}_{{k_i}}},{{\bf{a}}_{{k_i}}}} \right) = r + \gamma Q^C\left( {{{\bf{s}}_{{k_i} + 1}},\mu \left( {{{\bf{s}}_{{k_i} + 1}}|{\theta }} \right)}|{\phi^C} \right).
\end{align}

Then, the long-term reward and long-term reward cost neural networks are updated by using the mean-squared Bellman error (MSBE) loss function which are derived as follows
\begin{align}\label{17}
&L\left( \phi^R \right) = \\ \notag
& \quad \frac{1}{{{N_{\text{batch}}}}}\sum\limits_{i = 1}^{{N_{\text{batch}}}} { \left[ {{{\left( {{Q^R}\left( {{{\bf{s}}_{{k_i}}},{{\bf{a}}_{{k_i}}}} \right) - Q^R\left( {{{\bf{s}}_{{k_i}}},{{\bf{a}}_{{k_i}}}|\phi^R} \right)} \right)}^2}} \right]},
\end{align}

\begin{align}\label{18}
&L\left(\phi^C \right) =  \\ \notag
& \quad \frac{1}{{{N_{\text{batch}}}}}\sum\limits_{i = 1}^{{N_{\text{batch}}}} {\left[ {{{\left( {{Q^C}\left( {{{\bf{s}}_{{k_i}}},{{\bf{a}}_{{k_i}}}} \right) - Q^C\left( {{{\bf{s}}_{{k_i}}},{{\bf{a}}_{{k_i}}}|\phi^C} \right)} \right)}^2}} \right]}.
\end{align}
In primal-dual DDPG, the actor policy is updated by maximizing the Lagrangian function in (\ref{lagrangian}), where the long-term reward and the long-term cost are replaced by the critic network and the cost network, respectively, i.e.,

\begin{align}\label{19}
\mathop {\max }\limits_{{\theta } } \mathbb E\left[ {Q^R\left( {{{\bf{s}}_{{k_i}}},{{\bf{a}}_{{k_i}}}} \right) - \lambda {Q^C}\left( {{{\bf{s}}_{{k_i}}},{{\bf{a}}_{{k_i}}}} \right)} \right].
\end{align}

After that, the dual variable $\lambda$ is updated by gradient descent to minimize the Lagrangian function according to

\begin{align}\label{20}
{\lambda ^{(k + 1)}} = {\left[ {\lambda _i^{(k)} + {\beta _k}\left( {{Q^C}\left( {{{\bf{s}}_{{k_i}}},\mu \left( {{{\bf{s}}_{{k_i}}}|{\theta }} \right)}|{\phi^C} \right) - {\frac{{{\Gamma _c}}}{{1 - \gamma }}}} \right)} \right]^ + },
\end{align}
where ${\beta _k}$ is the step size and ${\left[ x \right]^ + } = \max \left\{ {0,x} \right\}$.

\subsection{KC-TD3 Design}
The straightforward application of primal-dual DDPG in our problem can cause several issues (to be discussed in the following). To improve the performance in the training process of primal-dual DDPG, we propose to exploit advanced reinforcement learning techniques and expert knowledge on sampling, communication, and prediction including 1) extension of double Q-Learning, 2) state-space reduction, 3) interdependent action normalization, and 4) \gls{apdo}. The resulting approach is KC-TD3. 
\subsubsection{Extension of Double Q-Learning}
The overestimation bias of the critic network will lead to poor performance when optimizing the actor network~\cite{fujimoto2018addressing}. Similarly, with primal-dual DDPG, if the cost network is underestimated, the constraint cannot be satisfied. In our problem, we have a constraint on the average synchronization error between the virtual model and real-world device. As in double Q-learning critic networks are trained to approximate the state-action value function, where the target value of the Bellman equation is the smaller one of the two critic networks. Thus, the Bellman equation can be expressed as follows,
\begin{align}\label{15}
Q^R\left( {{{\bf{s}}_{{k_i}}},{{\bf{a}}_{{k_i}}}} \right) = r + \gamma \mathop {\min }\limits_{l = 1,2} Q_{_{\phi _l^R}}^R\left( {{{\bf{s}}_{{k_i} + 1}},\mu \left( {{{\bf{s}}_{{k_i} + 1}}|{\theta}} \right)}|{\phi _l^R} \right),
\end{align}
where $Q_{_{\phi _l^R}}^R$ is the estimated state-action reward function from the $l$-th reward network with parameter ${{\phi _l^R}}$. Based on the knowledge of the synchronization error constraint that the synchronization error should be smaller than a required threshold, two cost networks are trained to approximate each state-action cost function. The target cost value is estimated by the larger one of the two cost networks. Thus, the Bellman equation can be expressed as follows,
\begin{align}\label{16}
{Q^C}\left( {{{\bf{s}}_{{k_i}}},{{\bf{a}}_{{k_i}}}} \right) = r + \gamma \mathop {\max }\limits_{l = 1,2} Q_{_{\phi _l^C}}^C\left( {{{\bf{s}}_{{k_i} + 1}},\mu \left( {{{\bf{s}}_{{k_i} + 1}}|{\theta}} \right)}|{\phi _l^C} \right),
\end{align}
where $Q_{_{\phi _l^C}}^C$ is the estimated state-action cost function of the $l$-th cost network with parameter ${{\phi _l^C}}$.

\subsubsection{State-Space Reduction}
The original state ${{\bf{s}}_k}$ consists of two trajectory segments measured by the device, ${\mathcal{T}}(k-1)$ and ${\mathcal{T}}(k)$, and that predicted by the cloud server, $\hat{\mathcal{T}}(k-1)$ and $\hat{\mathcal{T}}(k)$. According to the expert knowledge of the our prediction algorithm, the dimension of the state $\bf{s_k}$ is dynamic and depends on the action $W_k$. However, the dimension of the input of the actor network is fixed, and thus we cannot feed the state ${{\bf{s}}_k}$ to the actor network directly. One possible approach is to replace the trajectory segments with the input of the prediction algorithm, $\bar{\mathcal{T}}_{\rm in}$, which relies on the measured trajectory segments and determines the predicted trajectory segments. Although the dimension of $\bar{\mathcal{T}}_{\rm in}$ is fixed, it can be large when the correlation time of the trajectory is large (up to a few seconds) and the state generation rate is high ($1000$~samples/s in our prototype). This may lead to a long training time and require a large number of samples. To improve the learning efficiency, we replace the original state with the MSE in (6),
\begin{align}
{{\dot s}_k} = e\left( k \right).
\end{align}
In this way, we can reduce the input of the actor network to a scalar.

\subsubsection{Interdependent Action Normalization}
Considering the aforementioned action in Section \ref{Subsec:CMDP}, we design the action with two elements, ${{\bf{a}}_k}=[{{\bf{a}}_k}^{[1]},{{\bf{a}}_k}^{[2]}]$, where ${{\bf{a}}_k}^{[1]} \in \{D_{\text{max}}, ...,W_{\text{max}}\}$ and ${{\bf{a}}_k}^{[2]} = {1,...,{{\bf{a}}_k}^{[1]}}$. According to the knowledge that the sampling interval cannot exceed the length of one trajectory segment, the feasible region of the second element depends on the first element. Based on that we normalize ${{\bf{a}}_k}^{[2]}$ by using $W_k$ (inverse of ${{\bf{a}}_k}^{[1]}$), i.e.,
\begin{align}
\dot {{\bf{a}}}_k^{[2]} = \frac{{{\bf{a}}_k}^{[2]}}{{{W_k}}}.
\end{align}
With this normalization, we have $\dot {{\bf{a}}}_k^{[2]} \in [0,1]$ and use a $\rm sigmoid$ function in the output layer of the actor network.
\subsubsection{\gls{apdo}}
The dual variable updating procedure only utilizes on-policy samples, which leads to low sampling efficiency. To address this issue, we proposed to apply \gls{apdo}~\cite{liang2018accelerated}, where the dual variable is updated by the dual gradient ascent every $d_{\lambda}$ iteration (shown in the steps 16-18 in Algorithm~1). With this approach, the historical data samples stored in the replay buffer are utilized to update $\lambda$ and help to improve the sample efficiency and the convergent speed.

\begin{figure}
            \centering
            \includegraphics[scale=0.56]{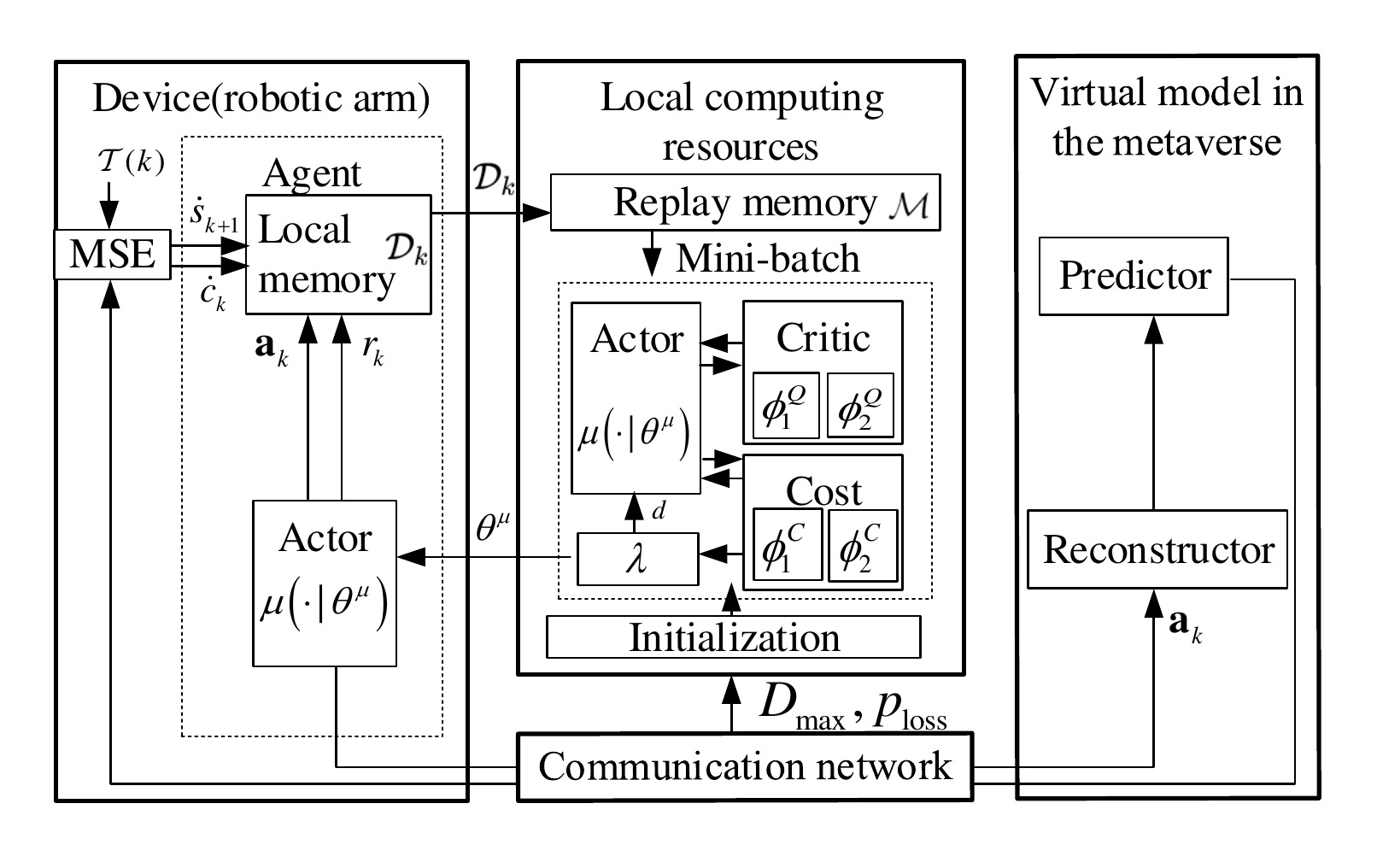}
          \caption{Illustration of proposed KC-TD3 architecture.}
          \label{Illustration of proposed TD3 architecture.}
  \end{figure}
\subsection{KC-TD3 Training Architecture}
The proposed KC-TD3 structure can be implemented in the real-world system with the architecture in Fig.~\ref{Illustration of proposed TD3 architecture.}, which mainly consists of a device (such as a robotic arm), local computing resource (a desktop/local server connected to the robotic arm), the communication network, and the metaverse. 

\subsubsection{Communication Network Initialization}
The delay, $D_{\max}$, and the packet loss probability, $p_{\text{loss}}$, in the communication network are measured in the initialization stage. 
With linear reconstruction in \eqref{reconstruct}, we need at least two samples to reconstruct each trajectory segment. Otherwise, the system is in outage. We denote the outage probability by $\o$. For a given requirement on the outage probability, such as $10^{-5}$, and a packet loss probability in the communication network (up to \SI{10}{\percent} in our experiments), the minimum number of samples the transmitter should update for each segment is denoted by $N_{\min}$. To meet the outage probability requirement, $N_{\min}$ can be obtained from the following expression,
\begin{align}\label{eq:outage}
{p_\text{loss}^{{N_{\min}}}} + {N_{\min}}{p_\text{loss}^{{N_{\min}} - 1}}(1 - p_\text{loss}) \le \o.
\end{align}

\subsubsection{Training Algorithm} The details of the algorithm can be found in Algorithm~1.
With the observed average tracking error, the agent at the device takes actions according to the output of the actor network. Then, the average tracking error of the next trajectory segment is measured and saved in the replay memory. After that, the transitions in the replay memory are randomly selected to optimize the critic and cost networks (referred to as one iteration of the gradient descent optimization). Finally, the actor network and the dual variable $\lambda$ are updated every $d_{\rm a}$ and $d_{\lambda}$ iterations (referred to as update delay in Algorithm~1), respectively.




\begin{algorithm}[t]\label{alg.l1}
\begin{algorithmic}[1]
\renewcommand{\algorithmicrequire}{\textbf{Input:}} 
\renewcommand{\algorithmicensure}{\textbf{Output:}} 

\caption{KC-TD3} 
\REQUIRE
Initialize parameters of actor, critic and cost networks, $\theta$, ${\phi _1^R}$, ${\phi _2^R}$, ${\phi _1^C}$, ${\phi _2^C}$. Measure the latency and packet loss probability in the communication network, $D_{\max}$ and $p_{\rm loss}$. Obtain the minimal number of samples the device needs to update for each trajectory segment from \eqref{eq:outage}. \\ 
\ENSURE
Optimal ${\lambda ^ * }$and optimal policy ${\mu ^ *(\cdot|\theta^{*})}$.

\STATE Initialize the target networks: ${\theta_{\arg}} \leftarrow \theta$, ${\phi^R _{\arg ,1}} \leftarrow {\phi ^R_1}$, ${\phi^R _{\arg ,2}} \leftarrow {\phi^R _2}$, ${\phi^C _{\arg ,1}} \leftarrow {\phi^C _2}$, ${\phi^C _{\arg ,2}} \leftarrow {\phi^C _2}$.
\STATE Initialize the Lagrangian multiplier $\lambda  = 0$, actor update delay $d_{\rm a}$, dual variable update delay $d_{\lambda}$.
\FOR{episode $m$ = 1,...} 
    \STATE Observe the average tracking error $\dot{s}_k$.
    \STATE Generate an action based on (\ref{14}) and execute the action.
    \STATE Observe the reward, cost, the next state, and store $\left\langle {{\dot s_{{k_i}}},{{\bf{a}}_{{k_i}}},{r_{{k_i}}},{c_{{k_i}}},{\dot s_{{k_i} + 1}}} \right\rangle$ in memory ${\rm{{\cal M}}}$.
        \FOR{$j$ in range (from $1$ to the maximal number of updates)}
            \STATE Randomly sample a batch of transitions $\left\langle {{\dot s_{{k_i}}},{{\bf{a}}_{{k_i}}},{r_{{k_i}}},{c_{{k_i}}},{\dot s_{{k_i} + 1}}} \right\rangle$ from ${\rm{{\cal M}}}$.
            \STATE Updating $Q^R$-functions by one step of gradient descent based on (\ref{15}), (\ref{17}).
            \STATE Updating $Q^C$-functions by one step of gradient descent based on (\ref{16}), (\ref{18}).
            \IF{$j$ mod $d_{\rm a}$  = 0}
                \STATE Update the actor policy by by one step of gradient ascent based on (\ref{19}).
            \ENDIF
            \IF{$k$ mod $d_{\lambda}$ = 0}
                \STATE Update dual variable by one step of gradient ascent based on (\ref{20}).
            \ENDIF
            \STATE Update target network by:
            \begin{align}
            {\phi^R _{\arg }} \leftarrow \rho {\phi^R _{\arg,i}} + (1 - \rho )\phi^R_i  \notag;\\
            {\phi^C _{\arg }} \leftarrow \rho {\phi^C _{\arg,i}} + (1 - \rho )\phi^C_i  \notag;\\
            {\theta_{\arg }} \leftarrow \rho {\theta_{\arg }} + (1 - \rho )\theta \notag;\\
            i = \{ 1,2\}.\notag 
            \end{align}
        \ENDFOR
\ENDFOR

\end{algorithmic}
\end{algorithm}

\section{Prototype Design and Data Collection}
\label{sec:experiment}

\subsection{Prototype Design}
\begin{figure}
            \centering
            \includegraphics[scale=0.25]{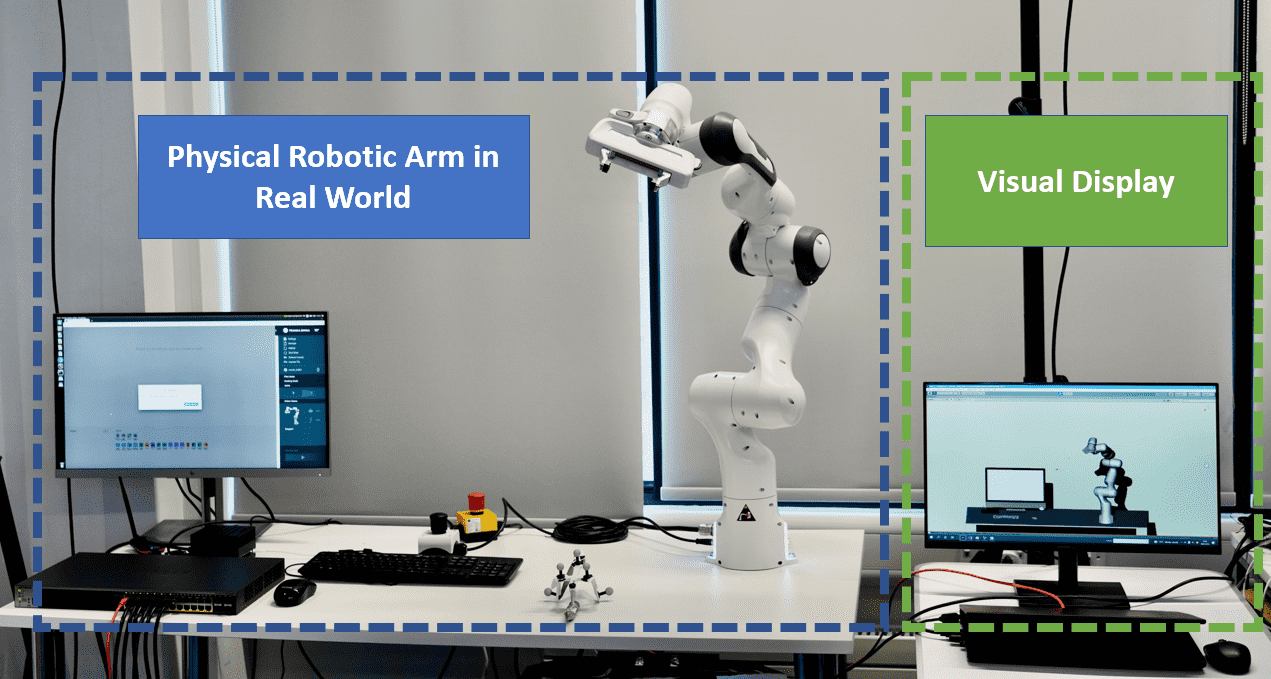}
           \caption{Our prototype system (the movements of a physical robotic arm and the visual display. The digital model in the metaverse to be synchronized is in the remote which is not shown in the graph).}
           \label{Motion-based real-time control system with digital twin}
  \end{figure}

We build a 
prototype\footnote{In our prototype, we simplify the system by considering the one-way synchronization from a single physical robotic arm to a virtual robotic arm, and test our co-design framework and KC-TD3 algorithm. The proposed co-design can be extended to multi-user cases.} as shown in Fig.~\ref{Motion-based real-time control system with digital twin}, where a virtual robotic arm needs to synchronize with a physical robotic arm in the real world. This is essential for many future use cases, such as education, healthcare, Industry 4.0, etc.

\textbf{Physical Robotic Arm in Real World:}
An industrial-grade robotic arm system, Franka Emika Panda~\cite{libfranka}, is used in our prototype. It has seven {degree of freedoms} (DoFs) achieving up to 2 m/s end-effector speed and +/-0.1 mm repeatability. The robotic arm receives the target end-effector position from a controller, and then conducts inverse-kinematics calculation to map the target end-effector position to the seven joint angle positions of the robotic arm. After that, the robotic arm applies a {proportional-integral-derivative} method~\cite{ang2005pid} for control, which converts the joint angle positions to a series of commands on the angular velocity of each joint. Multiple types of sensing data including joint angle values, joint angular velocities and inertial torque of joints are obtained by the application programming interface provided by libfranka, which is a C++ implementation of the Franka Control Interface~\cite{libfranka}.


\textbf{Virtual Robotic Arm in the Metaverse:}
Unity software is used to generate the virtual robotic arm in the metaverse~\cite{craighead2008using}. Specifically, we construct the digital model of the physical robotic arm, Franka Emika Panda, with the same number of DoFs deployed on a cloud server. The virtual robotic arm needs to obtain the angle position of each joint in real time. Such that the virtual robotic arm can be synchronized with the physical robotic arm. In the communication system, our prototype uses {User Datagram Protocol} to connect the virtual robotic arm to the physical one. Besides, the sensing data of the robotic arm are displayed to the user by High-Definition Screen via \gls{hdmi}.

\begin{figure}
            \centering
            \includegraphics[scale=0.27]{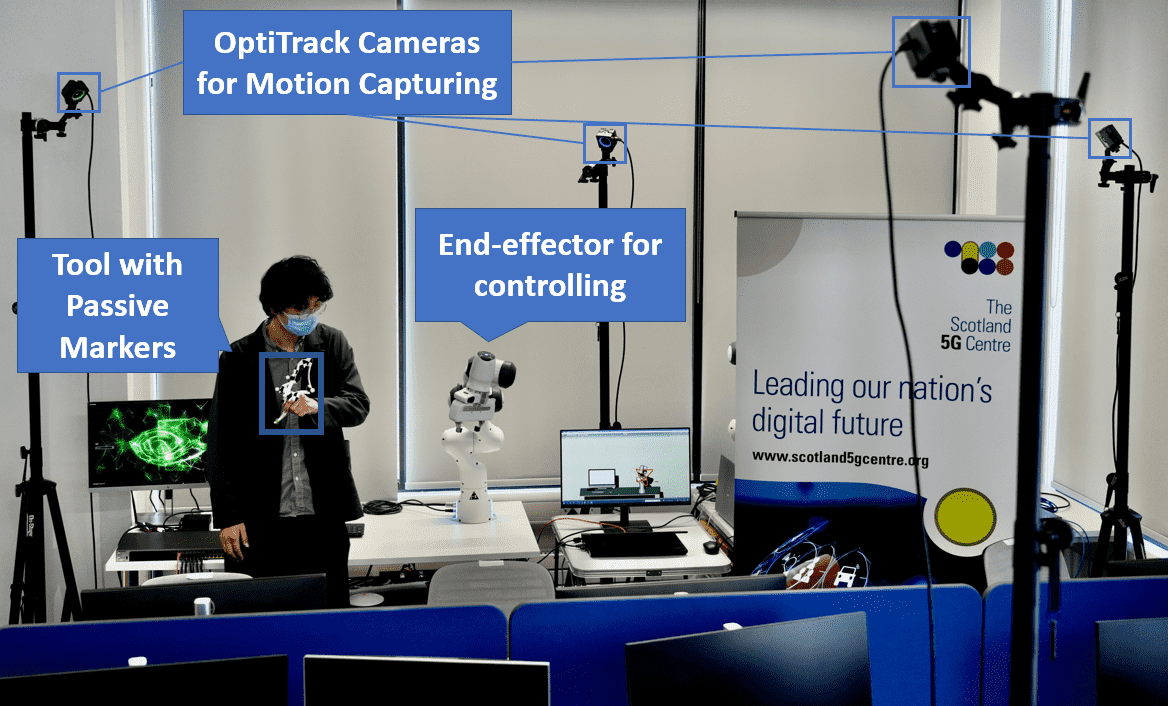}
          \caption{Illustration of our data collection via an experiment, where a human operator controls the physical robotic arm to draw the ``star'' shape in the air (The demonstration video of our data collection is available at \url{https://youtu.be/LCqSGtkrgug})}.
          \label{Robotic arm drawing shape '$star$' controlled by operator's hand motion}
  \end{figure}
  
  
\subsection{Data Collection}
As shown in Fig.~\ref{Robotic arm drawing shape '$star$' controlled by operator's hand motion}, a human operator controls the physical robotic arm via a motion capture system~\cite{optitrack}, drawing a ``star'' shape in the air for $20$ seconds.

Specifically, six OptiTrack Prime-13 motion capture cameras are deployed in a $4 \times 4\;\text{m}^2$ area, where the human operator holds a tool with seven passive markers~\cite{optitrack}. The motion capture system constructs the seven markers as a rigid body and outputs the position (3 DoFs) and orientation (3 DoFs) of the rigid body at the frequency of $120$~Hz. The robotic arm receives the position and orientation as the target position of its end-effector. By controlling the angles of the seven joints, the physical robotic arm is able to move its end-effector towards the target position. In this way, the end-effector of the robotic arm tracks the human operator's hand trajectory in real-time. In this work, we use the data of the first joint from the base to demonstrate our design. During the movement of the robot arm, the device measures the joint angles at a frequency of 1kHz, and saves the data in csv format files, which will be used in the subsequent neural network training\footnote{Unless otherwise specified, the settings of the experimental parameters are listed in TABLE I.}


\begin{table}[t]\label{parameter}
\renewcommand{\arraystretch}{1.5}
\centering
\caption{System Parameters for Performance Evaluation}
\begin{threeparttable} 
\begin{tabular}{|c|c|}
\hline 
\bf{Parameters} & \bf{Values}\\
\hline
Slot duration & \SI{1}{\milli\second} \\
\hline
Transmission time interval $T$ & \SI{1}{\milli\second} \\
\hline
Time delay bound $D_{\text{max}}$ & \SI{10}{\milli\second}\\
\hline 
The range of segment length $\traj(k)$ & \SIrange{10}{100}{\milli\second}\\
\hline 
The range of sampling rate $n(k)$ & \SIrange{20}{1000}{\hertz} \\
\hline
Experimental time & \SI{2e4}{\milli\second}\\
\hline 
\end{tabular}
\end{threeparttable}
\end{table}

\section{Performance evaluation}
\label{sec:results}

In this section, we first evaluate the training performance of our KC-TD3, and then compare the performance of the proposed sampling, communication, and prediction co-design framework with different benchmarks.

\subsection{Evaluation of KC-TD3 Algorithm}

\subsubsection{KC-TD3 Training Performance}

\begin{figure}[p]
  \centering
  \subfigure[Normalized average communication load in each episode (If the normalized average communication load is $100\%$, it means that the average communication load is the same as the system without sampling).]{
  \label{Reward}
  \includegraphics[scale=0.5]{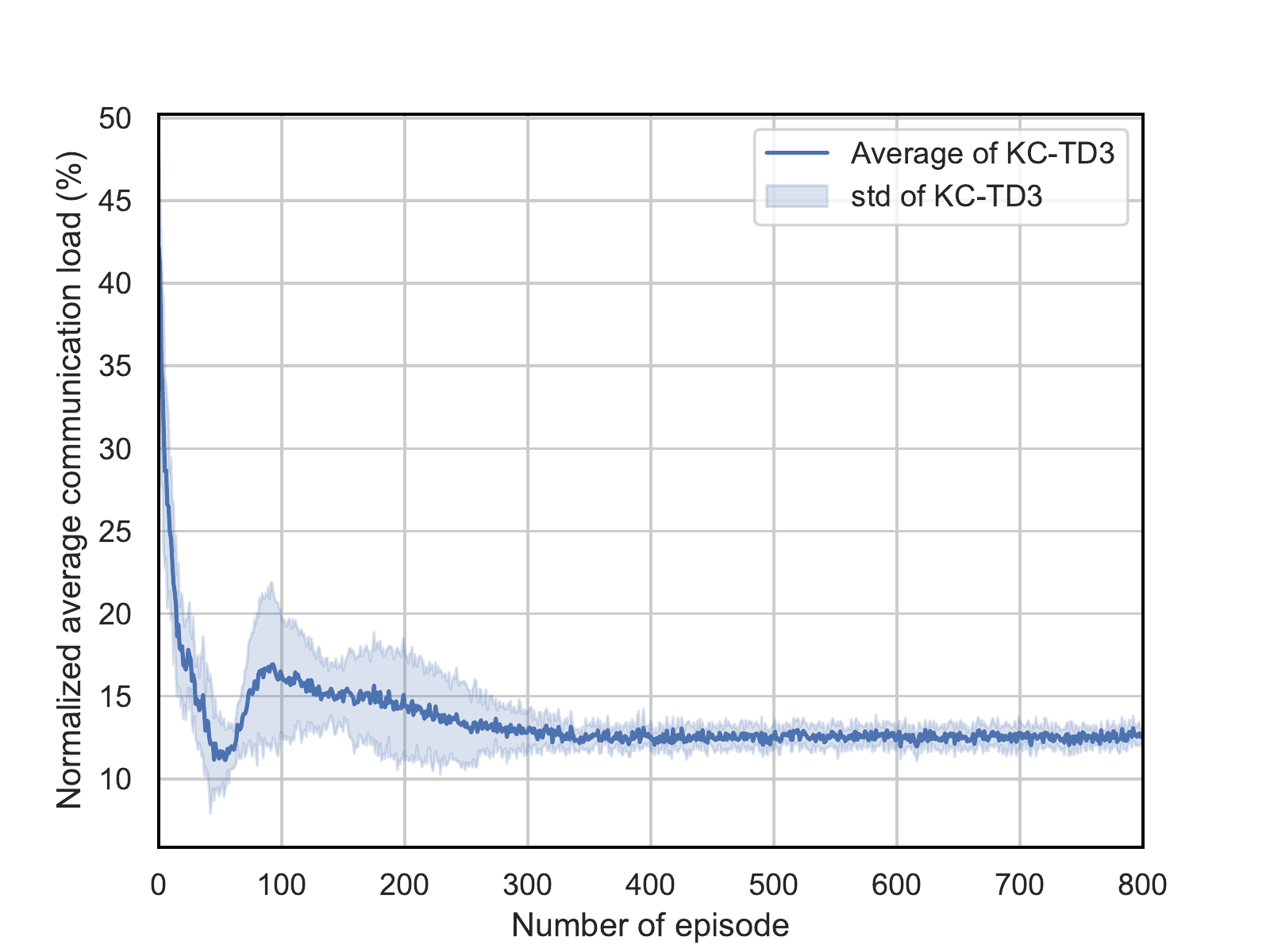}
  }

  \subfigure[Average tracking error in each episode.]{
  \label{Cost}
  \includegraphics[scale=0.5]{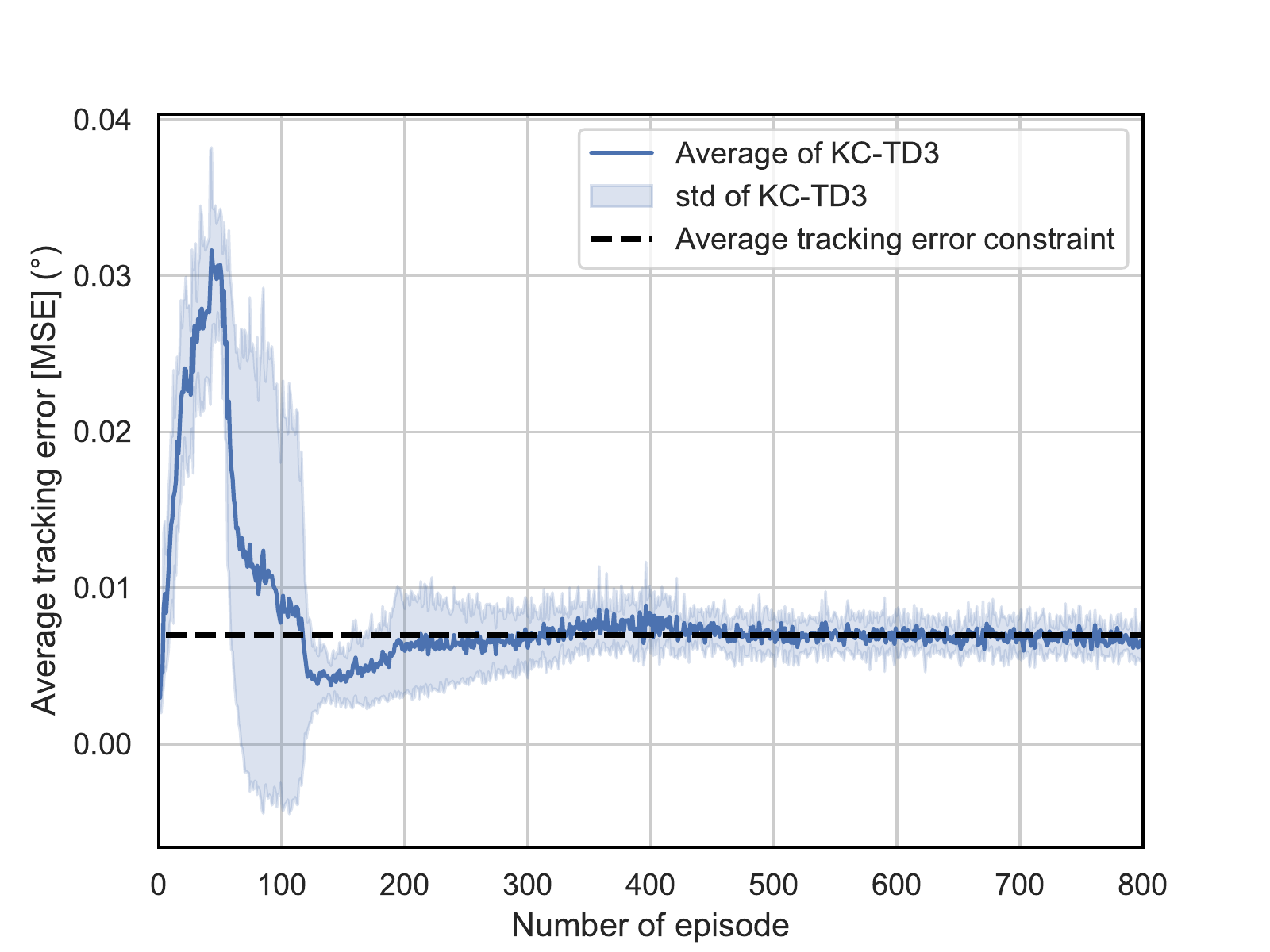}
  }

  \subfigure[Dual variable in each episode.]{
  \label{Lamda}
  \includegraphics[scale=0.5]{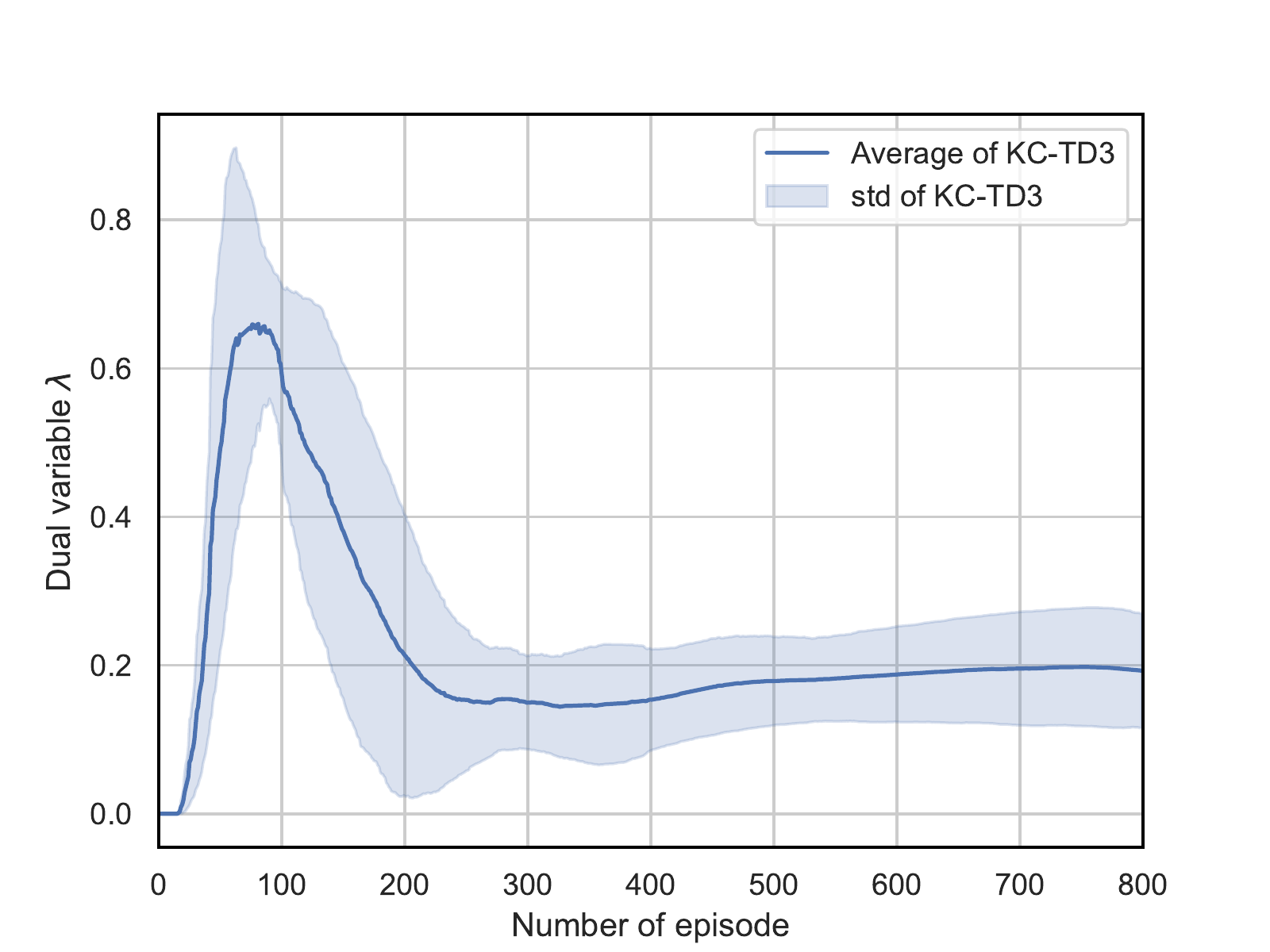}
  }  

  \caption{The average KC-TD3 training performance and standard deviation.}
  \label{fig:ENISA}
\end{figure}

The average KC-TD3 training performance and standard deviation are illustrated in Figs.~\ref{Reward}-\ref{Lamda}. In each training experiment, we provide the results in $800$ episodes. Then, we repeat the same experiment $10$ times. In Fig.~\ref{Reward}, we provide the normalized average communication load, where the communication load without sampling is \SI{100}{\percent}. In Figs.~\ref{Cost} and \ref{Lamda}, the average tracking error and the value of the dual variable are presented. The results in Figs. \ref{Reward}-\ref{Lamda}, show that the KC-TD3 algorithm converges to a stable policy after $300$ episodes. The normalized average communication load is $13\%$, and the average tracking error is $\text{MSE} = 0.007^{\circ}$, which satisfies the average tracking error constraint. 

\subsubsection{Ablation Study on Expert Knowledge} 
\begin{figure}[p]
  \centering
  \subfigure[Normalized average communication load in each episode.]{
  \label{domain_reward}
  \includegraphics[scale=0.5]{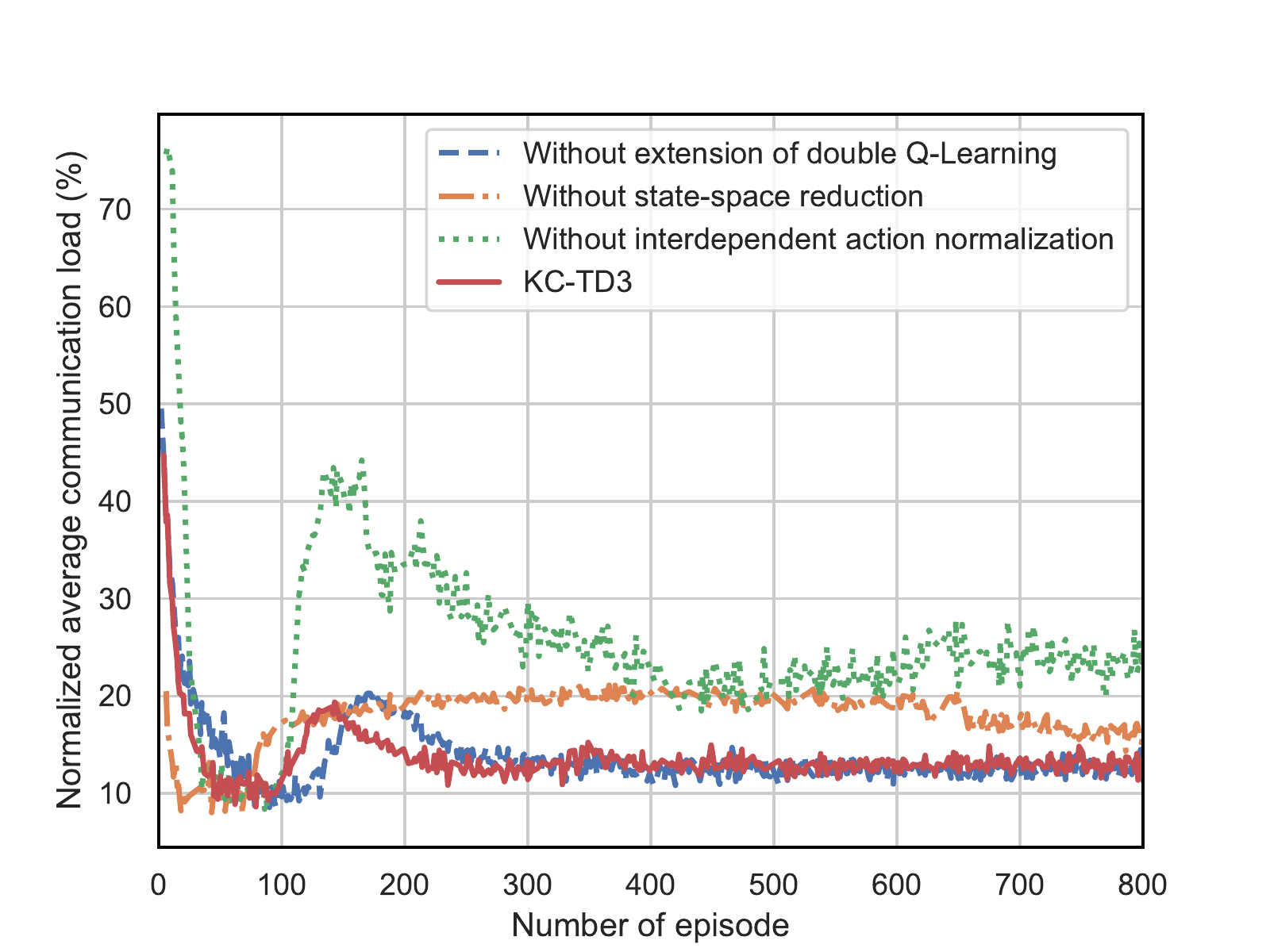}
  }

  \subfigure[Average tracking error in each episode.]{
  \label{domain_Cost}
  \includegraphics[scale=0.5]{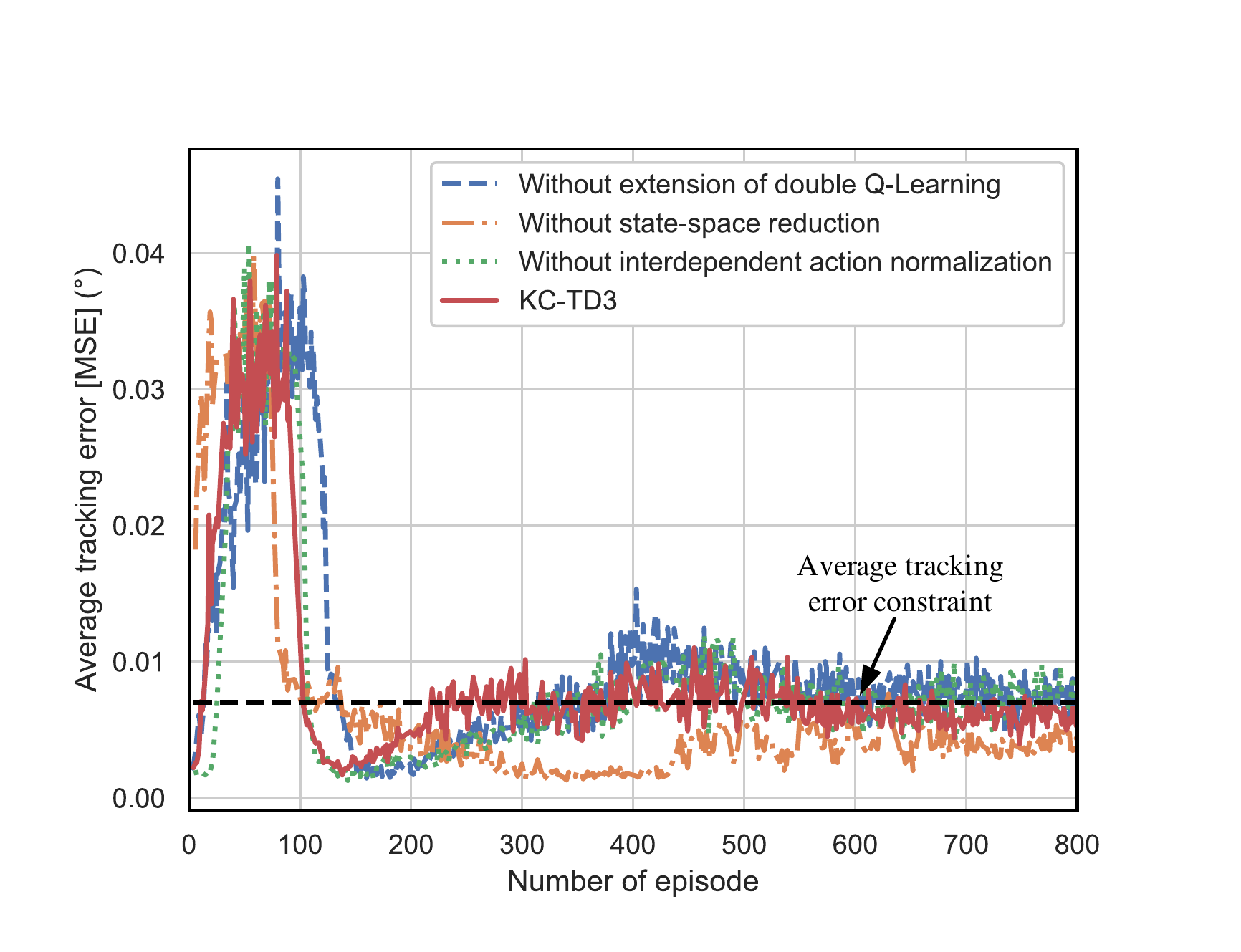}
  }

  \subfigure[Dual variable in each episode.]{
  \label{domain_Lamda}
  \includegraphics[scale=0.5]{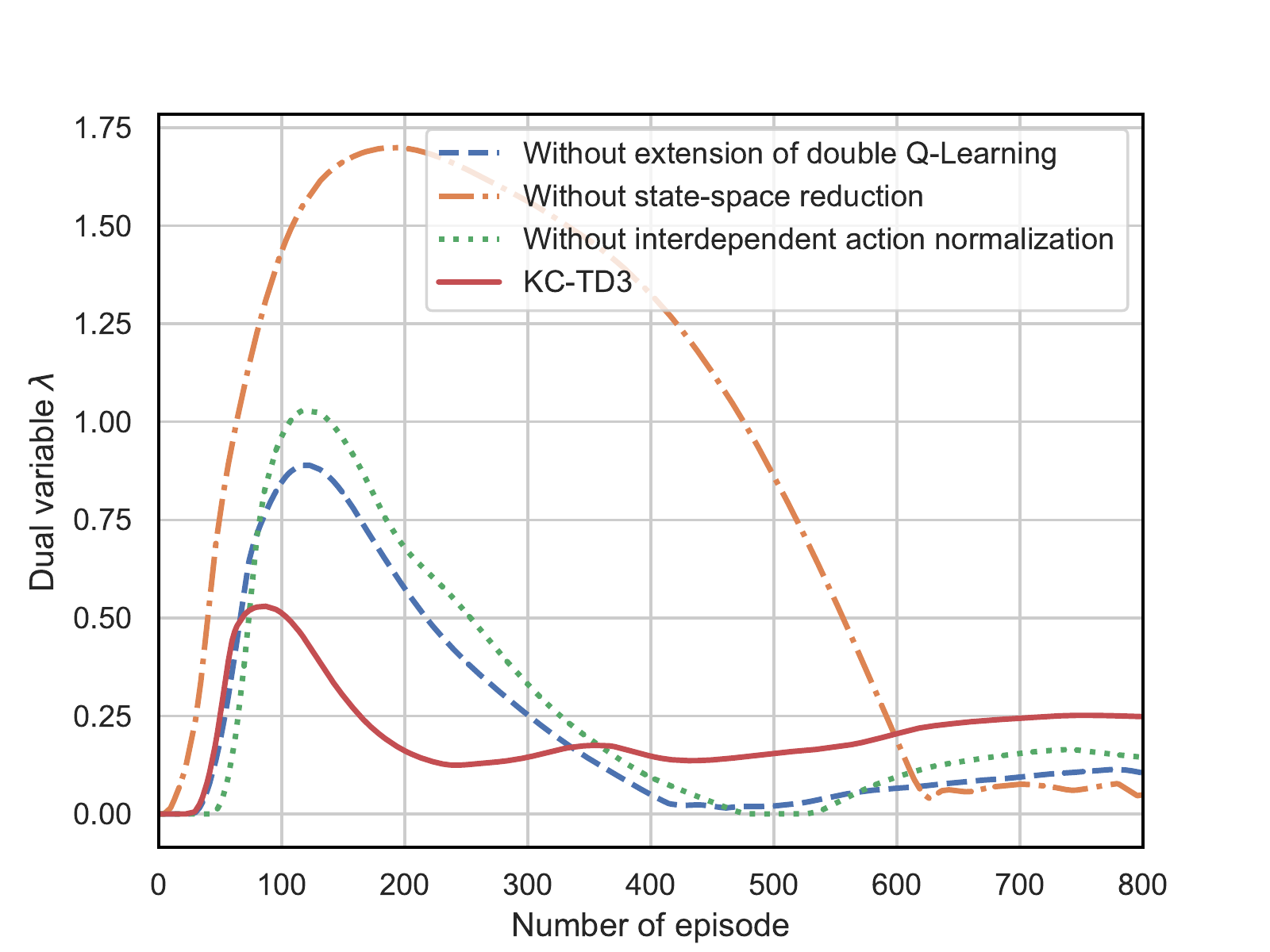}
  }  

  \caption{Ablation Study of different expert knowledge on the training performance of the proposed KC-TD3.}
  \label{fig:ENISA}
\end{figure}

In Figs.~\ref{domain_reward}-\ref{domain_Lamda}, we illustrate the impacts of different expert knowledge on the training performance of the proposed KC-TD3. When only partial expert knowledge is available, we obtained three benchmarks: (a) KC-TD3 without extension of double Q-Learning; (b) KC-TD3 without state-space reduction; (c) KC-TD3 without interdependent action normalization.

  



From Figs. \ref{domain_reward}-\ref{domain_Lamda}, we observe that the proposed strategy with full expert knowledge has the best performance in terms of stability and convergence. It also achieves the lowest communication load (in Fig.~\ref{domain_reward}), and satisfies the average tracking error constraint (in Fig.~\ref{domain_Cost}). By comparing the proposed KC-TD3 algorithm with the three benchmarks, we can obtain the following insights: (1) without the extended double Q-Learning, the algorithm can hardly meet the average tracking error constraint; (2) without state-space reduction, the algorithm cannot converge to the optimal solution, where the average tracking error is far below the constraint at the cost of a higher communication load. (3) Without the interdependent action normalization, the obtained policy can meet the average tracking error constraint, but it does not minimize the communication load.     

\subsection{Validation of the Sampling, Communication, and Prediction Co-design Framework}

\subsubsection{Fixed Sampling Policy}
\begin{figure}
            \centering
            \includegraphics[scale=0.5]{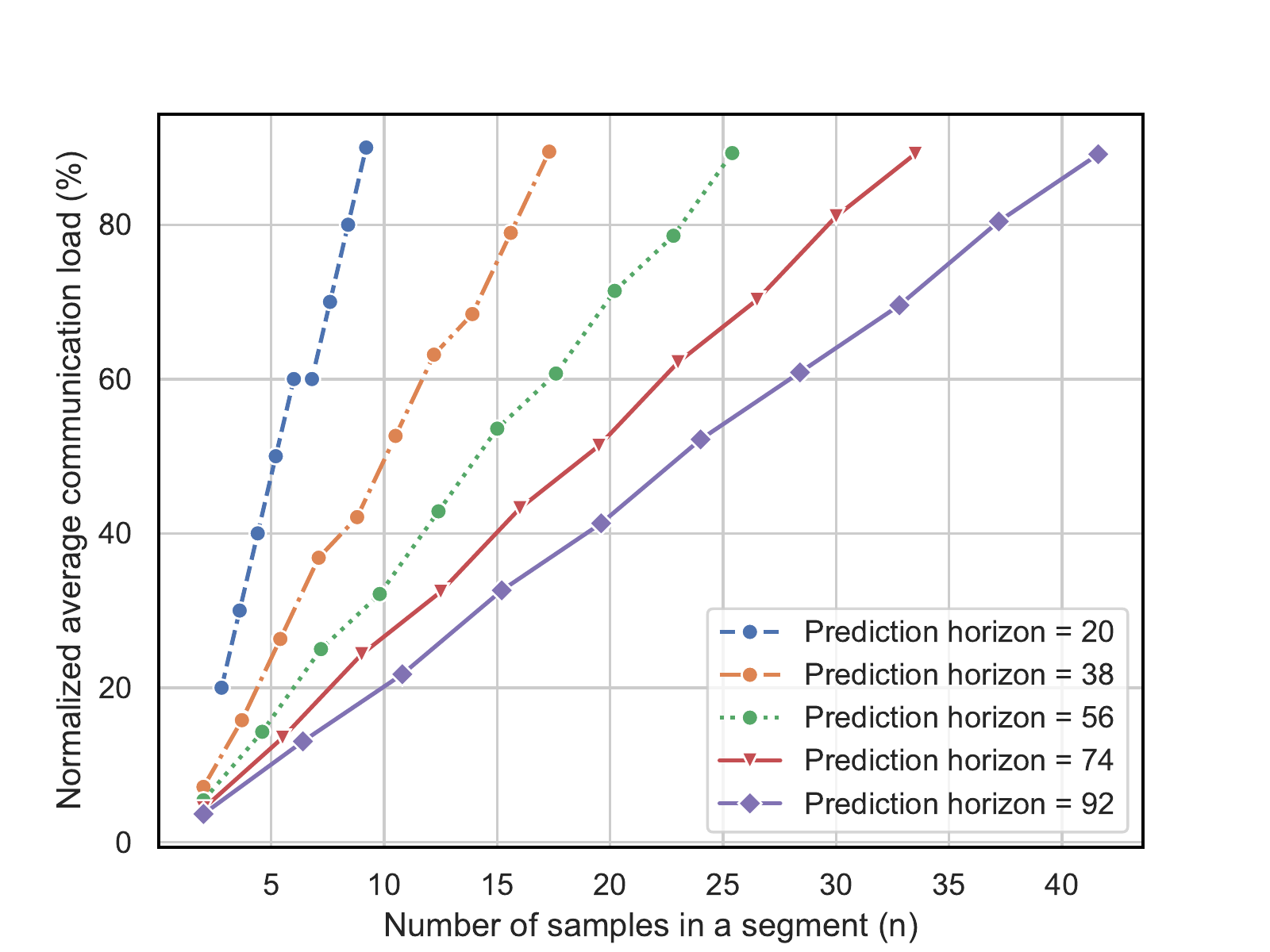}
           \caption{Normalized average communication load under different prediction horizons and sampling numbers.}
           \label{effiency}
  \end{figure}

\begin{figure}
            \centering
            \includegraphics[scale=0.5]{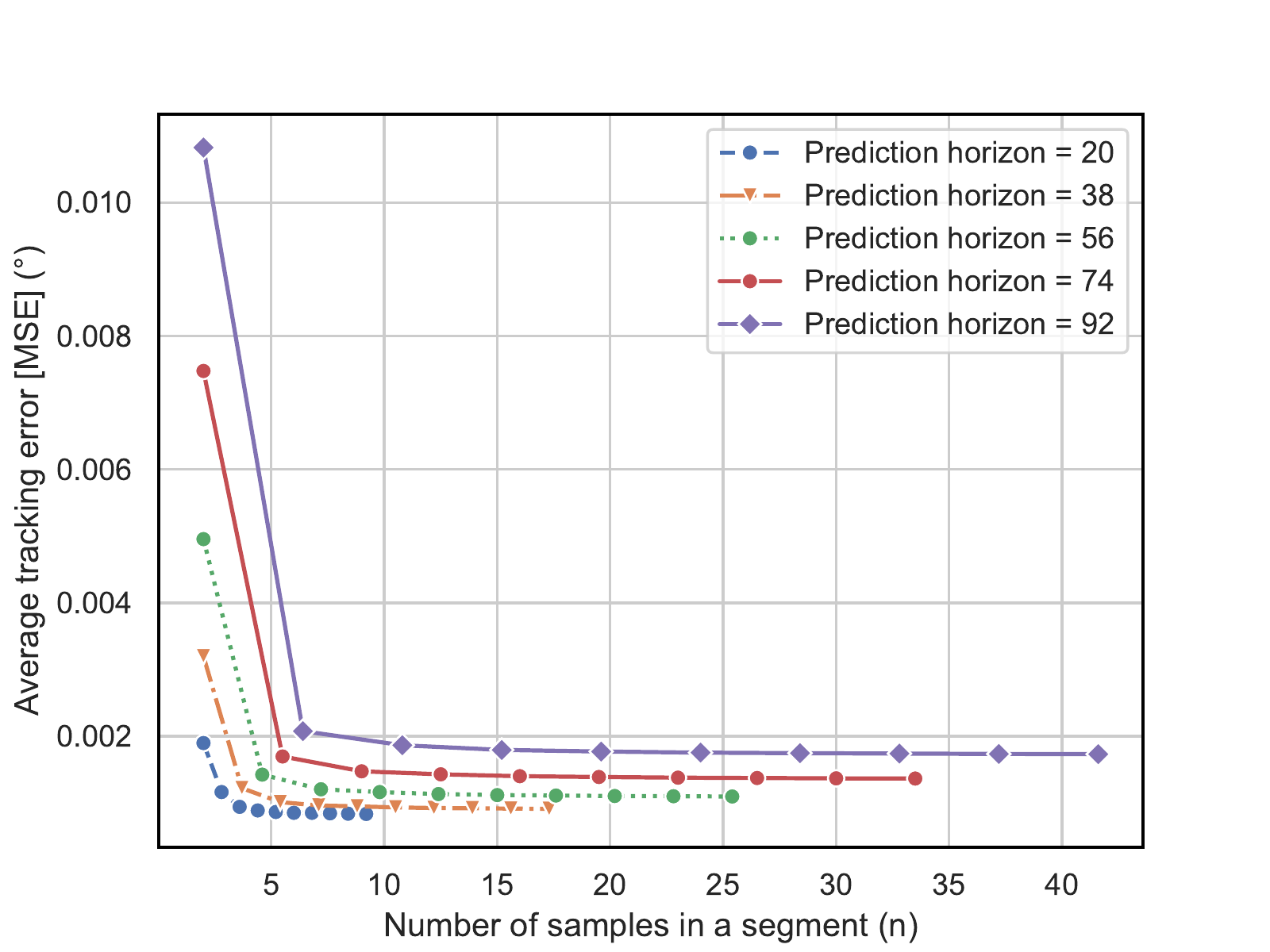}
           \caption{Average tracking error under different prediction horizons and the number of samples.}
           \label{average traking error}
  \end{figure}

We first evaluate the performance of a benchmark policy with fixed sample rate and prediction horizon. Specifically, Fig.~\ref{effiency} shows the normalized average communication load versus the number of samples in each segment. The results show that the normalized average communication load grows as the number of samples in each segment increases, but decreases as the prediction horizon increases. This means that a higher sampling rate (the ratio of the number of samples in each segment to the duration of the segment, which is half of the prediction horizon) leads to a higher communication load.


We further provide the average tracking errors with different sampling rates in Fig.~\ref{average traking error}. The results show that the average tracking error decreases dramatically as the number of samples in each segment $n$ increases from $2$ to $10$. When $n>10$, the average tracking error is nearly constant. In addition, a longer prediction horizon leads to a higher average tracking error. Therefore, the number of samples in each segment and the segment length should be optimized to obtain the minimum normalized average communication load subject to the average tracking error constraint.     



\begin{figure}
            \centering
           \includegraphics[scale=0.4]{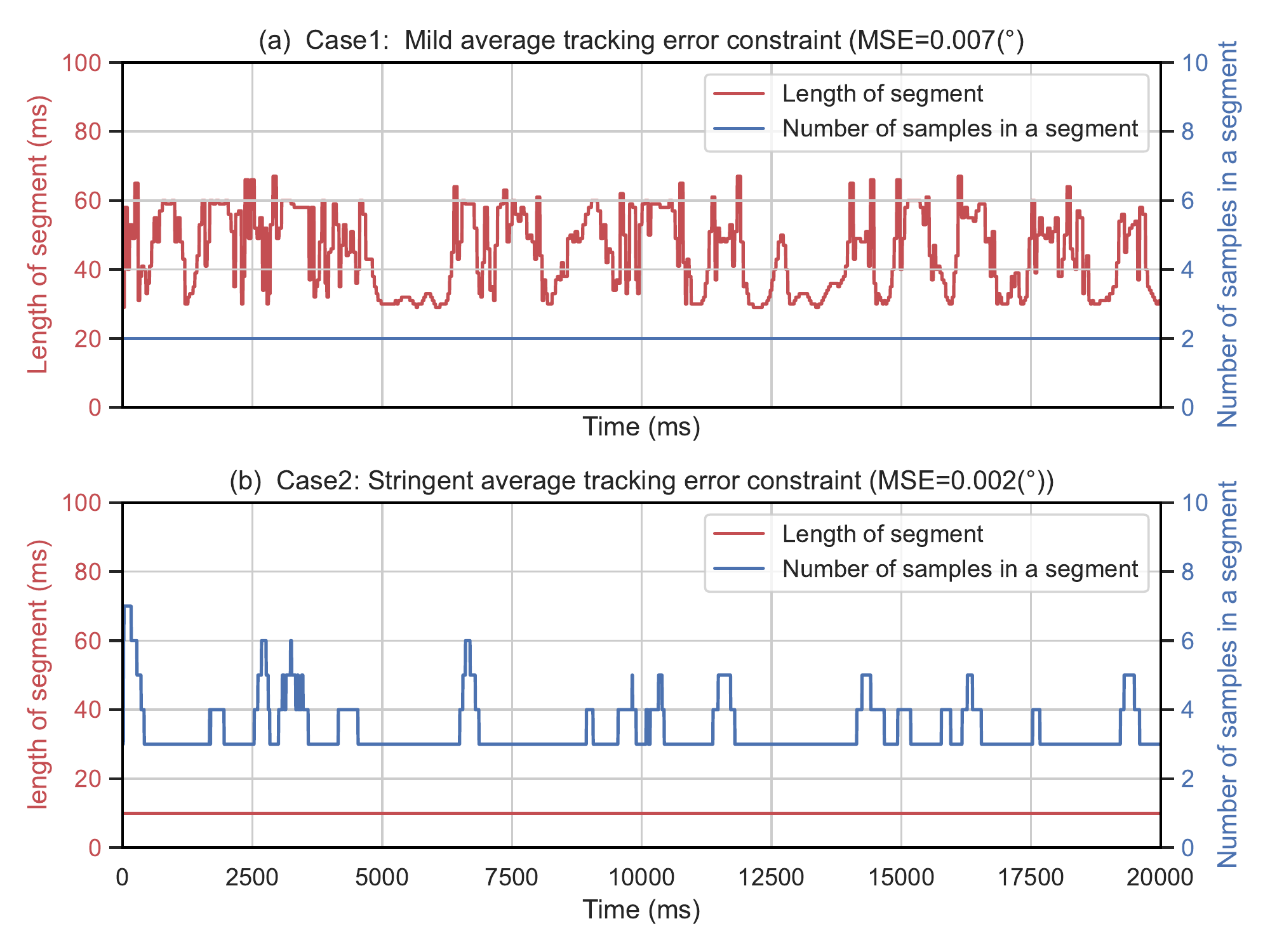}
           \caption{A demonstration of dynamic sampling of the proposed KC-TD3.}
           \label{Action taken in different tracking error constraints}
\end{figure}
  
\subsubsection{Dynamic Sampling in KC-TD3 }  
We provide an example in Fig.~\ref{Action taken in different tracking error constraints} to illustrate how the proposed KC-TD3 dynamically adjusts the segment length (which is equivalent to adjusting the length of prediction horizons) and the number of samples in each segment. Compared with existing approaches: (1) When the average tracking error constraint is stringent ($\text{MSE}=0.002^{\circ}$), our policy degenerates into the policy obtained from the sampling-communication co-design framework. (2) When the average tracking error constraint is mild ($\text{MSE}=0.007^{\circ}$), our policy degenerates into the policy obtained from the prediction-communication co-design framework. This indicates that our proposed strategy is a more general design framework compared with the two existing frameworks in the existing work. The results also indicate that the prediction-communication co-design framework cannot find the optimal policy when the tracking error constraint is stringent ($\text{MSE}=0.002^{\circ}$), and the sampling-communication co-design framework cannot find the optimal policy when the average tracking error constraint is mild ($\text{MSE}=0.007^{\circ}$).

\subsubsection{Overall Performance}

\begin{figure}
            \centering
            \includegraphics[scale=0.5]{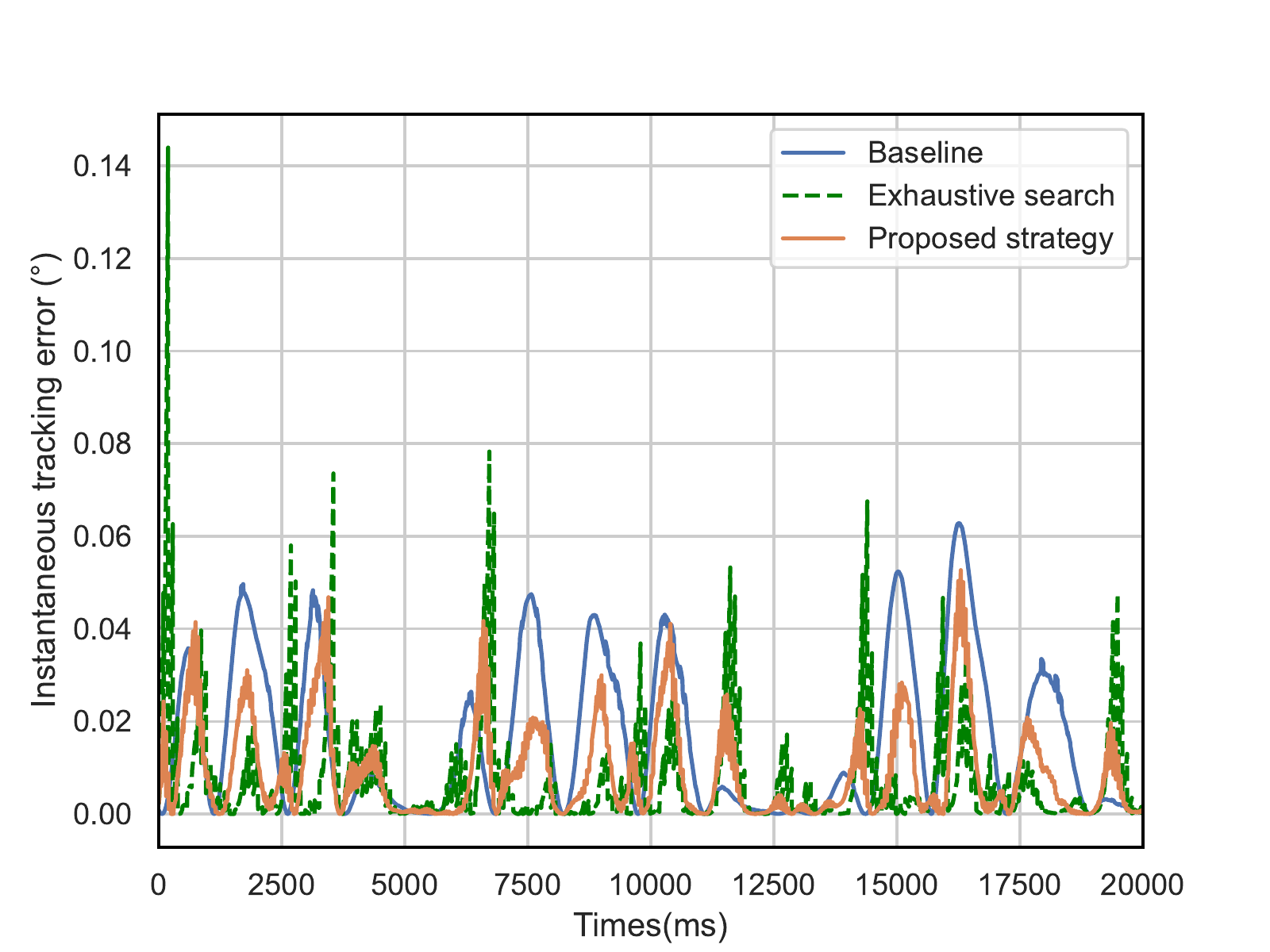}
           \caption{Instantaneous tracking error of Baseline, exhaustive   search (with optimized static $n_k$ and $H(k)$), and proposed KC-TD3 (with    optimized dynamic $n_k$ and $H(k)$), where the E2E latency is 50~ms and the average MSE constraint is $0.007^{\circ}$.}
           \label{gain}
  \end{figure}
  
\begin{figure}
            \centering
            \includegraphics[scale=0.5]{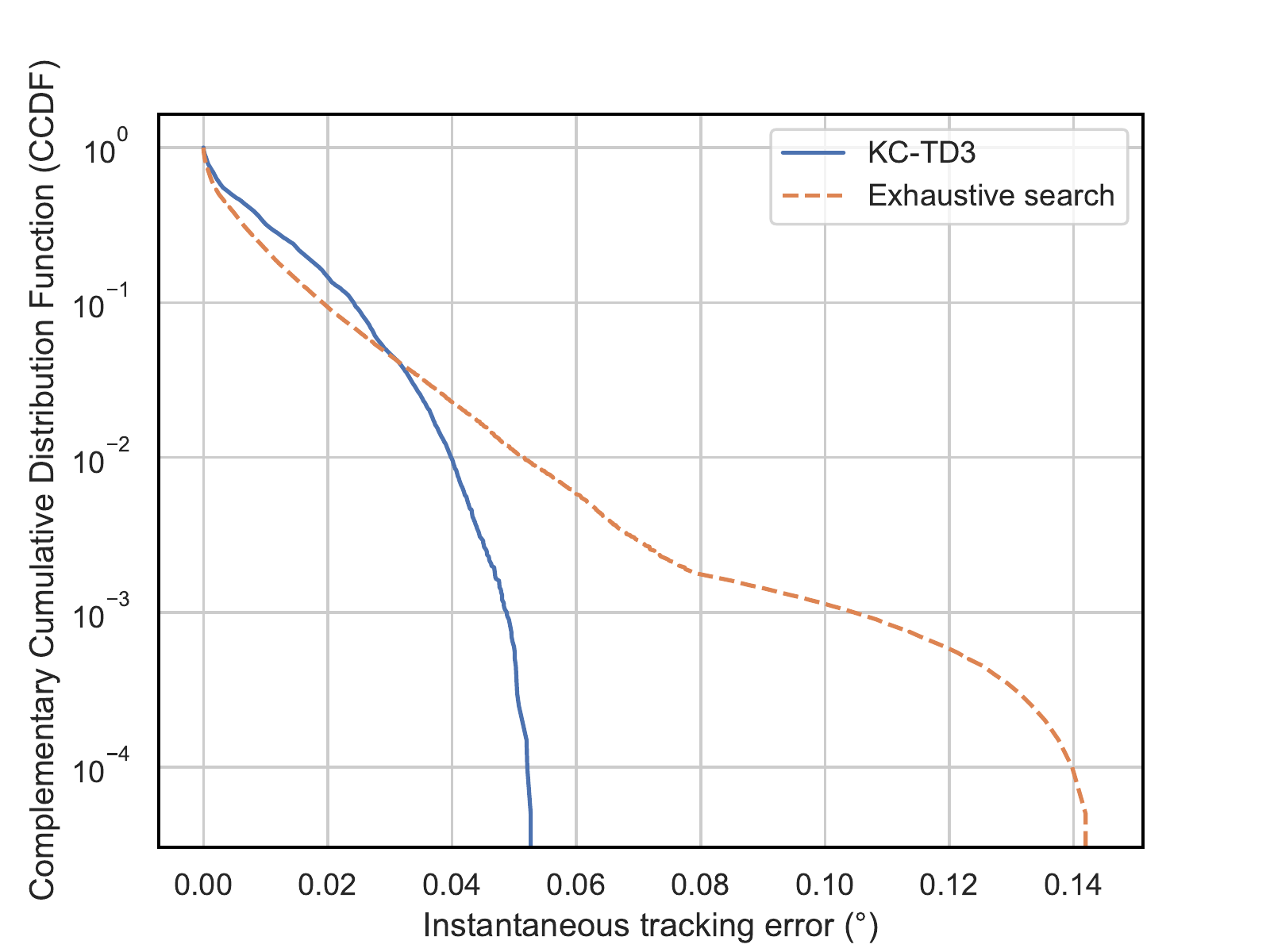}
           \caption{CCDF comparison of exhaustive search and proposed KC-TD3, where the E2E latency is 50~ms and the average MSE constraint is $0.007^{\circ}$.}
           \label{ccdf}
  \end{figure}
  
\begin{table}[]
\renewcommand{\arraystretch}{1.5}
\centering
\caption{PERFORMANCE COMPARISON OF DIFFERENT DESIGN}
\begin{tabular}{|cccc|}
\hline
\multicolumn{1}{|c|}{}                           & \multicolumn{1}{c|}{Baseline} & \multicolumn{1}{c|}{\begin{tabular}[c]{@{}c@{}}Exhaustive \\ search\end{tabular}} & KC-TD3 \\ \hline
\multicolumn{4}{|c|}{Average tracking error constraint = \SI{0.002}{\degree}}                                                                                                               \\ \hline
\multicolumn{1}{|c|}{\begin{tabular}[c]{@{}c@{}}Normalized average \\ communication load (\%)\end{tabular}} & \multicolumn{1}{c|}{100\%}    & \multicolumn{1}{c|}{27\%}                                                         & 27\%   \\ \hline
\multicolumn{1}{|c|}{Average tracking error (\SI{}{\degree})}     & \multicolumn{1}{c|}{0.016}    & \multicolumn{1}{c|}{0.0020}                                                       & 0.0019 \\ \hline
\multicolumn{4}{|c|}{Average tracking error constraint = \SI{0.007}{\degree}}                                                                                                               \\ \hline
\multicolumn{1}{|c|}{\begin{tabular}[c]{@{}c@{}}Normalized average \\ communication load (\%)\end{tabular}} & \multicolumn{1}{c|}{100\%}    & \multicolumn{1}{c|}{13\%}                                                         & 13\%   \\ \hline
\multicolumn{1}{|c|}{Average tracking error (\SI{}{\degree})}     & \multicolumn{1}{c|}{0.016}   & \multicolumn{1}{c|}{0.0070}                                                        & 0.0069  \\ \hline
\end{tabular}
\end{table}

Fig.~\ref{gain} compares the tracking errors of different design approaches, where the delay bound is $D_{\text{max}} = 50$ ms and the average tracking error constraint is MSE = $0.007 ^{\circ}$. The ``Baseline" approach transmits all samples to the receiver with the communication load of $100\%$ and there is no prediction at the receiver side. The ``Exhaustive search" approach is obtained by searching a fixed sampling rate and a fixed prediction horizon that minimize the normalized average communication load subject to the constraint on the average tracking error. We propose it here only as a performance benchmark, since it is not feasible for a practical online application. The instantaneous tracking errors achieved by the proposed strategy and the above two approaches are provided in Fig.~\ref{gain}, which shows that the proposed KC-TD3 can significantly reduce the instantaneous tracking error. This means that the synchronization between the virtual robotic arm and the physical one can be effectively improved. 

Table II compares KC-TD3 with ``Baseline" and ``Exhaustive search" approaches. Compared with ``Baseline" approach, KC-TD3 reduces the normalized average communication load by $73\%$ and improves the average tracking error by $87.5\%$ (when the average tracking error constraint is $0.002^{\circ}$). When the average tracking error constraint is $0.007^{\circ}$, KC-TD3 reduces the normalized average communication load by $87\%$ and improves the average tracking error by $56.2\%$ compared with ``Baseline" approach. Besides, the normalized average communication load and the average tracking error achieved by KC-TD3 are the same as ``Exhaustive search" approach.

Furthermore, we compared our policy that dynamically adjust sampling rate and prediction horizon according to the MSE with the exhaustive search approach that optimizes a static sampling rate and a static prediction horizon. We demonstrate the Complementary Cumulative Distribution Function (CCDF) of the tracking error in Fig.~\ref{ccdf}. The results indicate that the CCDF achieved by ``Exhaustive search" has a longer tail compared with the CCDF achieved by KC-TD3. This means that by adjusting the sampling rate and the prediction horizon dynamically, KC-TD3 can effectively reduce the tail probability of the tracking error. In other words, KC-TD3 enjoys more stable tracking error performance.


\begin{figure}
            \centering
            \includegraphics[scale=0.5]{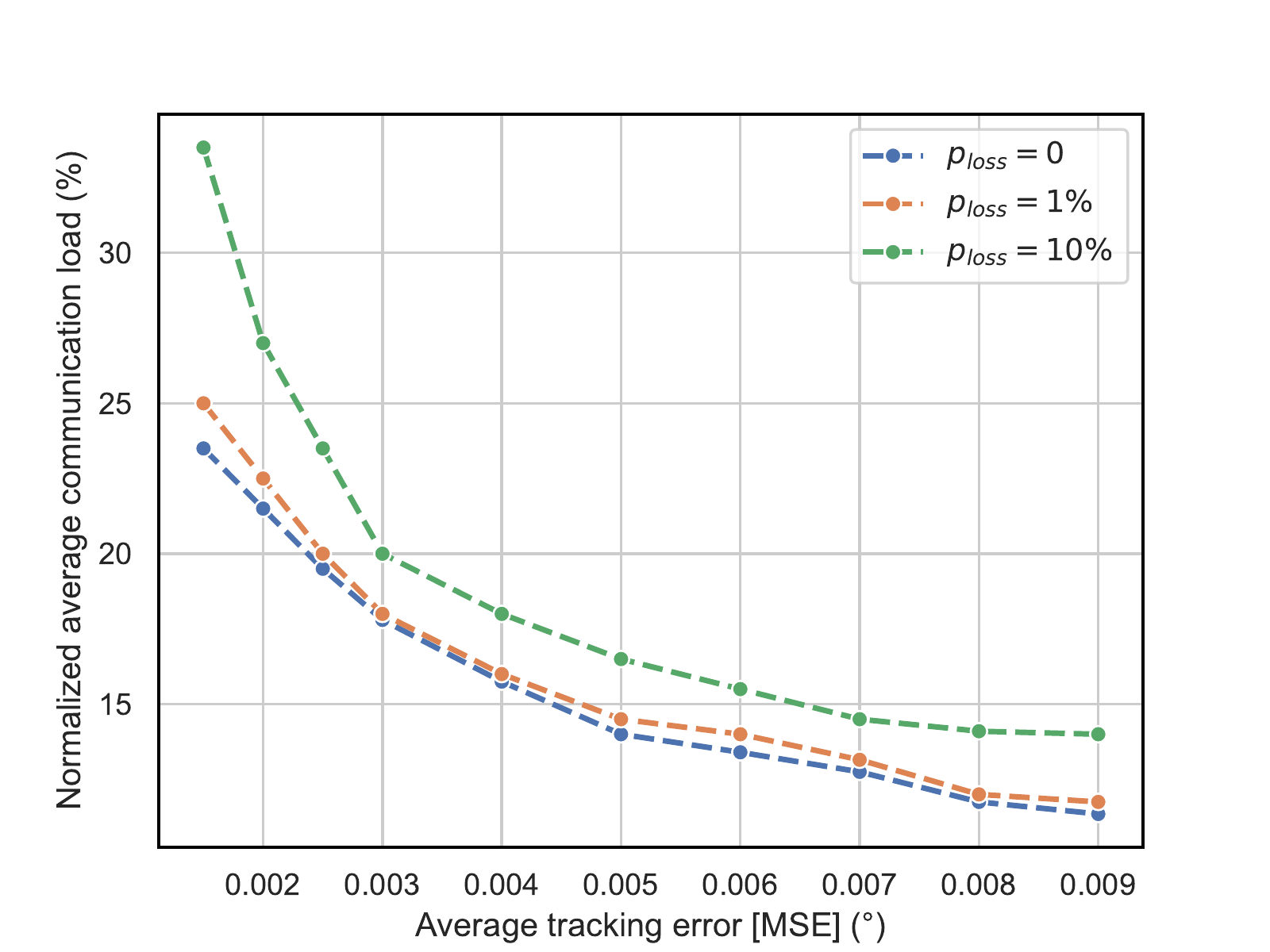}
           \caption{Trade-off between normalized average communication load and average tracking error with different packet loss probabilities $p_\text{{loss}}$ = 0, 1\% and 10\%.}
           \label{Fig_loss}
  \end{figure} 

Fig.~\ref{Fig_loss} further demonstrates the trade-off between the normalized average communication load and the average tracking error achieved by KC-TD3, where different packet loss probabilities in the communication system are considered, i.e.,  $p_\text{{loss}} = 0, 1\%,\ \text{and} \ 10\%$. The results reveal trade-off between the normalized average communication load and the average tracking error. In addition, with a smaller packet loss probability, it is possible to achieve a better trade-off between the normalized average communication load and the average tracking error. Furthermore, compared with ``Baseline" approach, KC-TD3 can reduce up to $75\%$ of normalized average communication load subject to a $0.002^{\circ}$ average tracking error constraint even when the packet loss probability in the communication system is as high as $10\%$.

        



\section{Conclusions}
\label{sec:conclusions}

In this paper, we demonstrated how to synchronize a real-world robotic arm and its digital model in the metaverse by sampling, communication, and prediction co-design. We established a framework for minimizing the average communication load under the constraint on the average tracking error between a real-world robotic arm and its digital model. Then, we proposed the KC-TD3 algorithm to adjust the sampling rate and the prediction horizon, where expert knowledge and advanced reinforcement learning techniques are exploited. In addition, we built a prototype of the proposed real-time robotic control system with a digital model in the metaverse. The results of our experiments showed that the proposed co-design framework can significantly reduce the communication load in the practical scenarios with communication packet losses. Compared with several benchmarks, our KC-TD3 algorithm converges faster and can guarantee the average tracking error constraint.

\begin{appendices}
\section{PROOF OF THE MARKOV PROPERTY}
\begin{figure}
            \centering
            \includegraphics[scale=0.61]{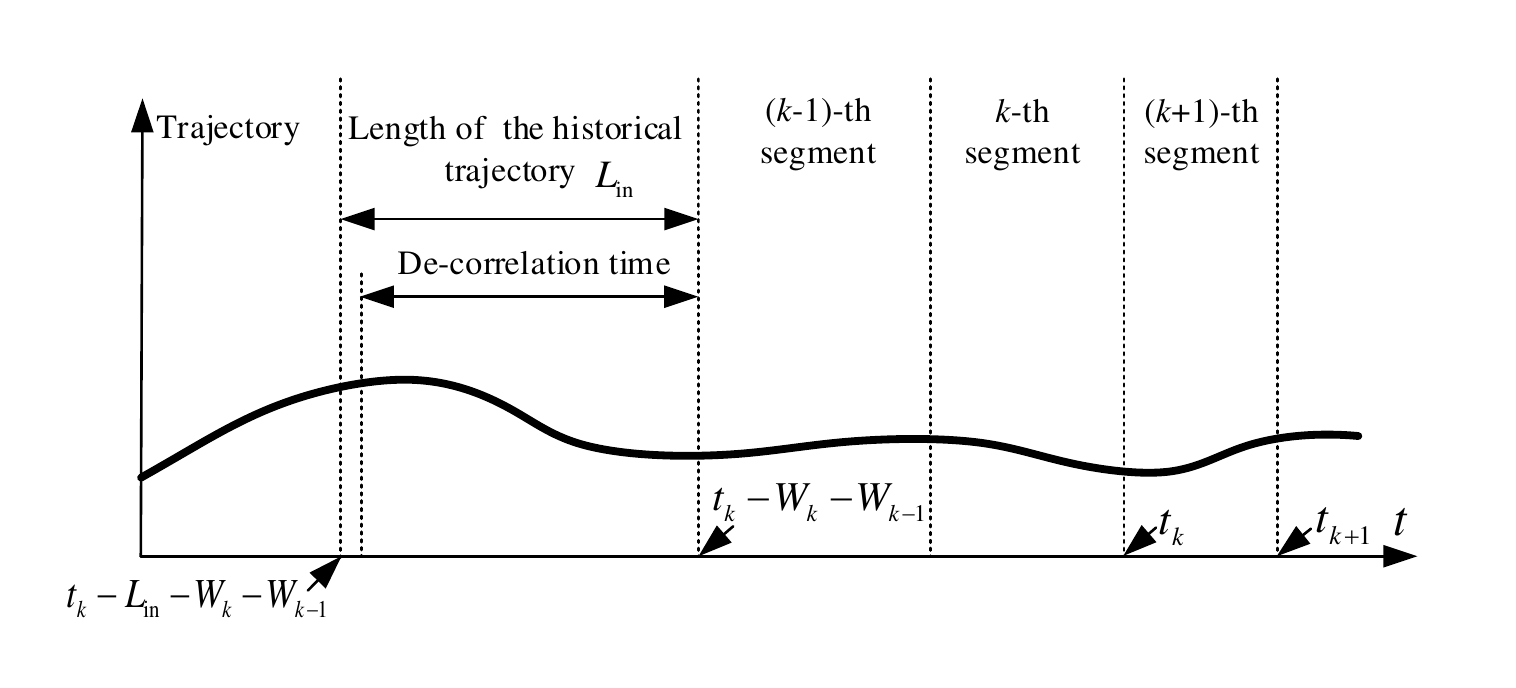}
           \caption{The relationship between the de-correlation time and length of the historical trajectory.}
           \label{de-correlation time}
  \end{figure} 
According to the definition, the $k$-th state includes, ${\mathcal{T}}(k-1)$, ${\mathcal{T}}(k)$, $\hat{\mathcal{T}}(k-1)$ and $\hat{\mathcal{T}}(k)$. As shown in \eqref{predict}, the predicted trajectory segments, $\hat{\mathcal{T}}(k-1)$ and $\hat{\mathcal{T}}(k)$, depend on the reconstructed historical trajectory, i.e., 
\begin{align}
 &\bar{\mathcal{T}}_{\rm in}(k) \nonumber\\
 &= [\bar{\tau}(t_{k}-L_{\rm in} -W_{k}-W_{k-1}),..., \bar{\tau}(t_{k}-W_{k}-W_{k-1})].\nonumber 
\end{align}
Further considering sampling and reconstruction in \eqref{eq:sample} and \eqref{reconstruct}, $\bar{\mathcal{T}}_{in}(k)$ is determined by the actions and trajectory from the $(t_{k}-L_{\rm in} -W_{k}-W_{k-1})$-th slot to the $(t_{k}-W_{k}-W_{k-1})$-th slot. Therefore,  ${\mathcal{T}}(k-1)$ and ${\mathcal{T}}(k)$ in the $k$-th state are determined by the states and actions in the past $(t_{k}-L_{\rm in} -W_{k}-W_{k-1})$ slots, which are available at the transmitter (the agent).

Similarly, the $(k+1)$-th state depends on the states and actions in the past $(t_{k+1}-L_{\rm in} -W_{k+1}-W_{k})$ slots, i.e., which is highly overlapped with that from the $(t_{k}-L_{\rm in} -W_{k}-W_{k-1})$-th slot to the $t_k$-th slot. The new trajectory information by the end of the $(k+1)$-th state includes $[\tau(t_{k+1}-W_{k+1}), ..., \tau(t_{k+1})]$ and the $k$-th action is ${{\bf{a}}_k}$. The dimension of the input of the prediction algorithm, $L_{\rm in}$, is the de-correlation time of the trajectory. Thus, $[\tau(t_{k+1}-W_{k+1}), ..., \tau(t_{k+1}))]$ only depends on the trajectory $[\tau(t_{k}-L_{\rm in} -W_{k}-W_{k-1}),...,\tau(t_{k})]$ (with the assumption $W_{k+1} < W_{k}+W_{k-1}$) \footnote{This assumption holds in most cases in our system since the prediction horizons of two consecutive segments are highly correlated and do not vary rapidly.} and action ${{\bf{a}}_k}$. In Fig.~\ref{de-correlation time}, we illustrates the relationship between de-correlation time and length of historical trajectory (observation horizon). Given the $k$-th state-action pair, if the observation horizon is longer than the de-correlation time of the system, the $(k+1)$-th state does not depend on the states and actions before the $(t_{k}-L_{\rm in} -W_{k}-W_{k-1})$-th slot. According to the definition of Markov decision process, the Markov property holds in our problem.

\end{appendices}
\normalem

\bibliography{IEEEabrv,bib}
\bibliographystyle{IEEEtran}
\vspace{-0.6in}

\ifthenelse{\boolean{review}}{
\clearpage
\newpage

\clearpage
\newpage

\clearpage
\newpage

}{}

\end{document}